%% file: main.tex
\documentclass{article} % For LaTeX2e
\usepackage{iclr2021_conference,times}

\input{math_commands}

\input{config}

\title{HALMA: \underline{H}umanlike \underline{A}bstraction \underline{L}earning \underline{M}eets \underline{A}ffordance in Rapid Problem Solving}

\author{Sirui Xie,  Xiaojian Ma, Peiyu Yu, Yixin Zhu, Ying Nian Wu, Song-Chun Zhu\\
UCLA Center for Vision, Cognition, Learning and Autonomy
}

\newcommand{\ditem}[2]{#1\small{$\pm$}#2}
\newcommand{\dicez}{\vcenter{\hbox{\includegraphics[width=0.021\textwidth]{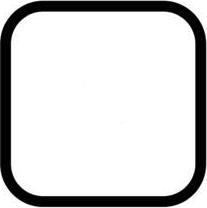}}}}
\newcommand{\diceone}{\vcenter{\hbox{\includegraphics[width=0.021\textwidth]{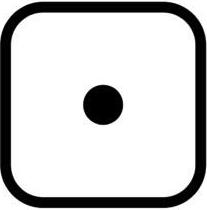}}}}
\newcommand{\dicetwo}{\vcenter{\hbox{\includegraphics[width=0.021\textwidth]{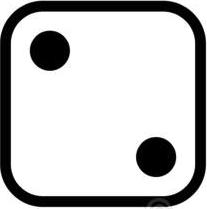}}}}
\newcommand{\dicethree}{\vcenter{\hbox{\includegraphics[width=0.021\textwidth]{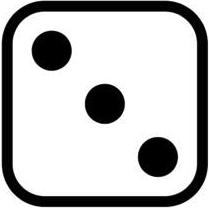}}}}
\newcommand{\Circle}{\vcenter{\hbox{\includegraphics[width=0.017\textwidth]{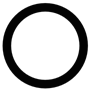}}}}
\newcommand{\Triangle}{\vcenter{\hbox{\includegraphics[width=0.017\textwidth]{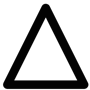}}}}
\newcommand{\Square}{\vcenter{\hbox{\includegraphics[width=0.017\textwidth]{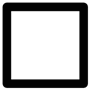}}}}
\newcommand{\Pentagon}{\vcenter{\hbox{\includegraphics[width=0.017\textwidth]{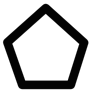}}}}
\newcommand{\caretu}{\vcenter{\hbox{\includegraphics[width=0.017\textwidth]{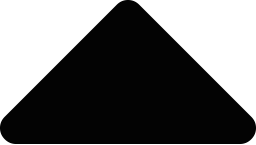}}}}
\newcommand{\caretd}{\vcenter{\hbox{\includegraphics[width=0.017\textwidth]{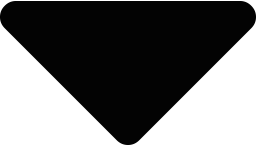}}}}
\newcommand{\caretl}{\vcenter{\hbox{\includegraphics[height=0.017\textwidth]{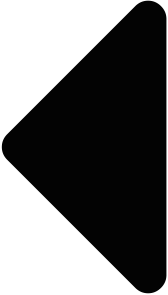}}}}
\newcommand{\caretr}{\vcenter{\hbox{\includegraphics[height=0.017\textwidth]{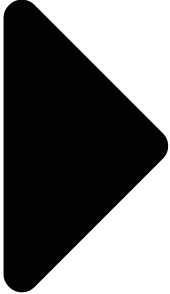}}}}
\newcommand{\mnist}[2]{\vcenter{\hbox{\includegraphics[width=0.022\textwidth]{mnist_dev/#1_#2}}}}

\newcommand{\stoptocwriting}{%
  \addtocontents{toc}{\protect\setcounter{tocdepth}{-5}}}
\newcommand{\resumetocwriting}{%
  \addtocontents{toc}{\protect\setcounter{tocdepth}{\arabic{tocdepth}}}}

\iclrfinalcopy % Uncomment for camera-ready version, but NOT for submission.
\begin{document}

\maketitle

\setstretch{0.98}

\begin{abstract}
Humans learn compositional and causal abstraction, \ie, knowledge, in response to the structure of naturalistic tasks. When presented with a problem-solving task involving some objects, toddlers would first interact with these objects to reckon what they are and what can be done with them. Leveraging these concepts, they could understand the internal structure of this task, without seeing all of the problem instances. Remarkably, they further build cognitively executable strategies to \emph{rapidly} solve novel problems. To empower a learning agent with similar capability, we argue there shall be three levels of generalization in how an agent represents its knowledge: perceptual, conceptual, and algorithmic. In this paper, we devise the very first systematic benchmark that offers joint evaluation covering all three levels. This benchmark is centered around a novel task domain, HALMA, for visual concept development and rapid problem solving. Uniquely, HALMA has a minimum yet complete concept space, upon which we introduce a novel paradigm to rigorously diagnose and dissect learning agents' capability in understanding and generalizing complex and structural concepts. We conduct extensive experiments on reinforcement learning agents with various inductive biases and carefully report their proficiency and weakness.\footnote{Check \url{https://halma-proj.github.io/} for a pilot of the HALMA environment. We will make HALMA and tested agents publicly accessible upon publication.}
\end{abstract}

\stoptocwriting
\section{Introduction}

\begin{wrapfigure}{Rt!}{0.44\linewidth}
    \vspace{-54pt}
    \centering
    \includegraphics[width=\linewidth]{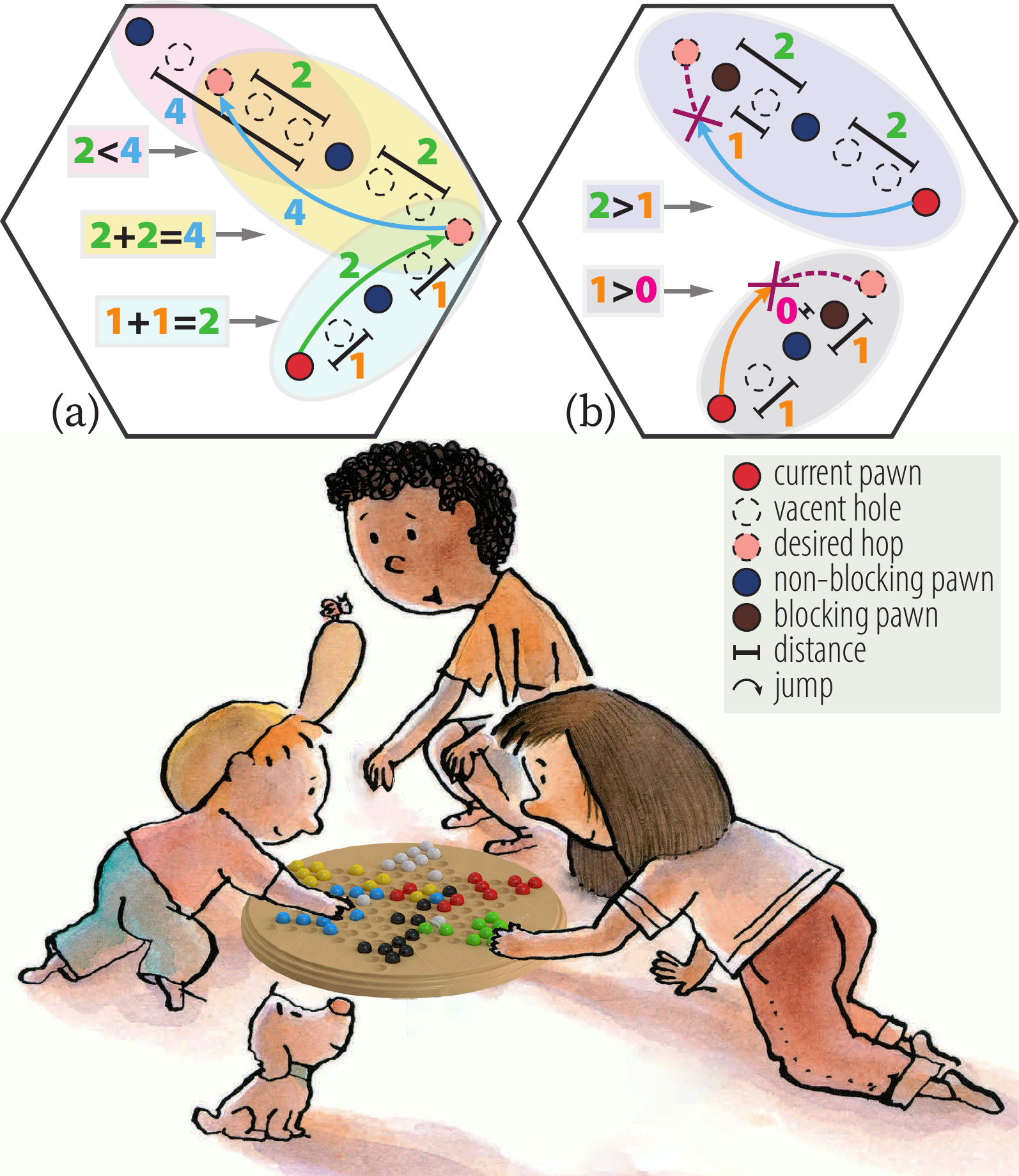}
    \caption{Illustration of the \emph{Super Halma} playing task. By playing the game with scarce supervision, Ada should be able to learn basic concepts of numbers and arithmetic, such as concepts with both (a) valid and (b) invalid moves (jumps).}
    \label{fig:halma_game}
    \vspace{-18pt}
\end{wrapfigure}

Have you ever heard of \emph{Super Halma},\footnote{See \url{https://en.wikipedia.org/wiki/Chinese_checkers\#Variants} for details.} a fast-paced variant of \emph{Halma}? In case you have not played \emph{Halma} or its fast-paced variant before, we briefly introduce both of them here. \emph{Halma} is a strategic board game, also known as \emph{Chinese checkers}. The rules of \emph{Halma} are minimal; it can be perspicuously explained using basic concepts of numbers and arithmetic. To win the game, one needs to transport pawns initially in one's own camp into the target camp. In each turn, a player could either \emph{move} into an empty adjacent hole and end the play, or \emph{jump} over an adjacent pawn, place on the opposite side of the jumped pawn, and \emph{recursively} apply this jump rule till the end of the play. While the standard rules allow hopping over only a single adjacent occupied position at a time, \emph{Super Halma} allows pieces to catapult over multiple adjacent occupied positions in a line when hopping; see an illustration in \cref{fig:halma_game}. We will use the term \emph{Halma} to specifically refer to \emph{Super Halma} in the remainder of the paper.

Now, imagine you are teaching your preschool cousin, Ada, to play \emph{Halma}. Since she has not yet formed a complete notion of \emph{natural numbers} or \emph{arithmetic}, verbally explaining the rules to her will render in vain. Alternatively, you can play with her while providing scarce supervisions, \eg, if a move is allowed; you can even reward her when she successfully moves a pawn to the target camp. By the time Ada could independently and rapidly solve unseen scenarios, we would know she has mastered the game. How many scenarios do you think Ada has to play before achieving this goal? 

This \emph{Halma} playing task is quintessential in the open-ended world; its environment is a minimal yet complete playground to test the rapid problem-solving capability of a learning agent. Under limited exposure to the underlying structure of the complex and immense concept space, we humans, by observing and interacting with entities, could form \emph{abstract} concepts of ``what it is'' and ``what can be done with it.'' The former one is dubbed \emph{semantics} \citep{jackendoff1983semantics} and the latter \emph{affordance} \citep{gibson1986ecological}. These abstract concepts, once accepted as knowledge, generalize robustly over scenarios; they are considered as milestones of human evolution in abstract reasoning and general problem solving \citep{holyoak1996mental}. In the case of \emph{Halma} playing task, Ada would be able to solve unseen scenarios \emph{within no time} if she were able to master (i) the abstract concept of natural numbers, emerged from and grounded to visual stimuli, (ii) both valid and invalid moves, and (iii) causal relations and potential outcomes risen from the grounded natural numbers and valid actions.

% When Ada learns to play \emph{Halma}, she needs to understand i) the similarity between distinct visual stimuli that allows the same primitive move, ii) why some stimuli allow her to plan for a teleportation composed with primitive moves, and iii) how to leverage these moves to reach the goal faster. She may finally come up with the grounded concept of natural numbers, at least within two digits. As such, she can solve new problems in \emph{Halma} within no time.
% , and master this game with the help of arithmetic

What is the proper machinery to learn these generalizable concepts from scarce supervisions?
By \emph{scarce supervision}, we mean the way to provide supervision is akin to how you teach Ada; one only provides sparse and indirect feedback without direct rules or dense annotations.
By \emph{generalizable concepts}, we emphasize more than the competence of memorization and interpolation; the learned representation ought to appropriately extrapolate and generalize in out-of-distribution scenarios. Such a superb generalization capability is often regarded as one of the celebrated signatures of human intelligence \citep{lake2015human,marcus2018algebraic,lake2018generalization}; it is attributed to rich \emph{compositional} and \emph{casual} structures in human mind \citep{fodor1988connectionism}.
% , and its emerging and developmental process has been systematically studied in developmental psychology \citep{carey2009origin}. 
% We have \emph{primitive concepts} extracted from perception such as the concept of objects developed in the first two months of our lives \citep{spelke2007core}. Around a year later we can syntactically combine them to form \emph{composed concepts} \citep{chomsky1957syntactic,montague1970universal}. At the age of 18 to 24 months, we can distinguish \emph{bootstrapped concepts} \citep{quine1960word} such as \emph{a walkable step is not a cliff} \citep{kretch2013cliff}.
Inspired by these observations, in this work, we quest for a computational framework to learn abstract concepts emerged in challenging and \emph{interactive} problem-solving tasks, with a humanlike generalization capability: The learned abstract knowledge should be easily transferred to out-of-distribution scenarios.

The general context of interactive problem solving poses extra challenges over classic settings of concept learning; instead of merely emerging concepts, it further demands the learning agent to leverage such emerged concepts for decision-making and planning. Ada, after understanding semantics and affordance in \emph{Halma}, can effortlessly perceive and parse novel scenarios \citep{zhu2020dark}. Yet, she would still struggle in strategically playing the game as she needs to decide among multiple affordable moves. In essence, the central question is: If conceptual knowledge can generalize as such, what meta-benefits does it offer on solving unseen problems \citep{schmidhuber1996simple}? 
% We argue that it at least comes with two major benefits. First, the concept of \emph{affordance} could improve the search efficiency by pruning the action space \citep{khetarpal2020can}; namely, leveraging task-specific abstraction can generically improve planning efficiency in sequential decision-making \citep{sanner2008first,jiang2015abstraction}. Second, one may realize that acting \emph{greedily} is not optimal; since \emph{Halma} is inherently \emph{partially observable} with unperceivable dead-ends, exploiting the longest affordable move in a new problem may not always be the best for exploration efficiency \citep{kaelbling1998planning}. Taking together, Ada may leverage these potential benefits based on learned concepts to discover a cognitively executable \emph{strategy} for rapid problem-solving in unseen scenarios. This strategy is what we call the \emph{algorithm} or \emph{heuristics} of this task.
The classic decision-making account of these meta-benefits would be: Leveraging knowledge, we can develop cognitively executable \emph{strategies} with high planning \citep{sanner2008first} and exploration efficiency \citep{kaelbling1998planning}; these strategies facilitate us to solve problems rapidly in unseen scenarios. They are what we call the \emph{algorithms} or \emph{heuristics} of this task.
% The meta-benefits discussed above shall, in turn, incentivize machines to learn abstract knowledge. Although 
Taking a step further, \citet{wang2018prefrontal,guez2019investigation} hypothesize that modern reinforcement learning agents, incentivized by these meta-benefits, have already discovered such algorithms. However, to date, their argument is still speculative since these agents have not been evaluated in tasks with rich internal structures yet limited exposure \citep{lake2017building,kansky2017schema}. A diagnosis benchmark for generalization capability is thus in demand to bridge communities of concept development and decision-making.
% their achievement is limited in memorizing and interpolating within seen scenarios; the methods are fairly restricted in extrapolation given a task with rich internal structures and limited exposure, likely for the reason that their model-agnostic principles neglect these structures in human mind \citep{lake2017building,kansky2017schema,dubey2018investigating}. The research program we propose here invites \emph{joint efforts} from communities of concept development and decision-making.

The main contribution of this paper is a \emph{Halma}-inspired competence benchmark: \ac{halma}. We rigorously devise \ac{halma} with three levels of generalization in visual concept development and rapid problem solving; see details in \cref{sec:three_levels}. \ac{halma} is unique in its \emph{minimum yet complete} concept spaces, a miniature of compositional and causal structures in human knowledge. It \emph{dynamically} generates test problems to informatively evaluate learning agents' capability in out-of-distribution scenarios \emph{under limited exposure}. We conduct extensive experiments with reinforcement learning agents to benchmark proficiency and weakness. 

\section{Three Levels of Generalization}
\label{sec:three_levels}

Our motivations might seem, \emph{prima facie}, bold. To convince readers and support our optimism, we summarize some recent progress in this section. In particular, we provide a taxonomy of three levels of generalization on a competency basis. Indeed, generalization is a multifaceted phenomenon. Previous evaluations for generalization were predominantly defined in a statistical sense, following the classical paradigm of train-evaluation-test random split \citep{cobbe2019quantifying} while ignoring internal structures. However, we argue this classical paradigm should not be the only objective approach wherein agents can or should generalize beyond their experience \citep{barrett2018measuring}, especially if our goal is to construct humanlike general-purpose problem-solving agents \citep{lake2017building}.
%We wish the taxonomy presented here can help motivate future investigations.

\setstretch{0.99}

\paragraph{Perceptual Generalization}
Perceptual generalization characterizes agents' capability to represent unseen perceptual signals, \eg, \emph{appearance} or \emph{geometry} in vision. In his seminal book, \emph{Vision}, \citet{marr1982vision} describes the process of vision as constructing a set of representations, parsing visual sensory data into descriptions. Such descriptions provide \emph{conceptual primitives} \citep{carey2009origin} for agents' understanding of the environment, boosting the efficacy of downstream cognitive activities (\eg, memory, learning, and reasoning). Learning an object-oriented representation of independent generative factors without supervision is thus believed to be a crucial precursor for the development of humanlike artificial intelligence. Although unsupervised disentanglement and segmentation \citep{eslami2016attend,higgins2016beta} resurged years ago, it is only till \citet{locatello2019challenging} did we realize the importance of evaluation on their generalization. More recently, \citet{burgess2019monet}, \citet{greff2019multi}, and \citet{lin2019space} evaluate their disentanglement/segmentation models outside of training regimes, especially on unseen combinations of visual attributes and numbers of objects. 

Although a hypothetically perfect \emph{semantic} description can truthfully represent the primitive concept of ``what it is,'' it could only contribute partially to achieving the understanding of ``what can be done with it'' \citep{montesano2008learning,zhu2015understanding}. Humanlike agents should equip with such task-oriented abstraction, \emph{affordance}, supported by compelling evidences in the field of developmental psychology; for instance, 18 to 24-month-old infants can distinguish \emph{bootstrapped concepts} \citep{quine1960word}, such as ``a walkable step is not a cliff'' \citep{kretch2013cliff}.
% Given a task specified by a primitive action and a primitive intent, irrelevant features to a task should be \emph{abstract out} \citep{khetarpal2020can}. When the task is specified by a Markov decision process, the abstraction is further desired to preserve the optimal value \citep{li2006towards,ferns2011bisimulation}; 
At a computational level, given a task specified by a Markov decision process, irrelevant features should be \emph{abstracted out} \citep{li2006towards,ferns2011bisimulation,khetarpal2020can}. Representation learned in this way bootstraps conceptual content. Recently, disentanglement as such has demonstrated efficacy \citep{gelada2019deepmdp,wayne2018unsupervised} and elementary perceptual generalizability \citep{zhang2020learning}. 

\paragraph{Conceptual Generalization}
While perceptual generalization closely interweaves with vision and control, conceptual generalization resides completely in cognition, assuming the readiness of all primitive concepts and some bootstrapped ones. The central challenge in conceptual generalization\footnote{Conventionally, it is dubbed \emph{combinatorial} generalization or \emph{systematic} generalization. We use the term \emph{conceptual} to highlight its functional signature.} is: How well can an agent perform in unseen scenarios given \emph{limited exposure} to the underlying \emph{configurations} \citep{grenander1993general}? It is connected with the Language of Thought Hypothesis \citep{fodor1988connectionism,goodman2008rational}: The productivity, systematicity, and inferential coherence in languages characterize compositional and causal generalization of concepts \citep{lake2015human}. 

How to learn representations with conceptual generalization is still an open question, drawing increasing attention in our community. With a synthetic translation task, \citet{lake2018generalization} reveal the incompetence of general purpose recurrent models \citep{elman1990finding,hochreiter1997long,chung2014empirical} in generalizing to (i) unseen primitives, (ii) unseen compositions, and (iii) longer sequences than training data. Similar incompetence of relational inductive biases \citep{battaglia2018relational} on hard compositional extrapolation has also been exemplified in abstract visual reasoning \citep{barrett2018measuring}. % \citet{lake2019compositional} further discusses the potentialities of some advanced inductive biases such as external memory, meta learning \citep{santoro2016meta} and attention \citep{vaswani2017attention} in extrapolating to unseen compositions, as well as their limitation to a close-ended domain of primitives. 
Notably, there is also a line of research on \emph{emerging} these linguistic structures from bootstrapped communication \citep{lazaridou2018emergence,mordatch2018emergence}. 

\paragraph{Algorithmic Generalization}
Agents' understanding of the structured environment should be reflected in their performance in solving novel problem instances; they ought to build strategies upon the developed concepts, resembling \emph{cognitive control} in human mind \citep{rougier2005prefrontal,botvinick2014computational}. We use the term algorithmic generalization to describe such flexibility. Specifically, for a problem domain where the internal structure contains an optimal exploration strategy, algorithmic generalization requires agents to discover this optimal strategy to explore efficiently in \emph{new} problem instances. For example, in the domain of dependent bandit problems designed by \citet{wang2016learning}, there is one arm whose return leaks the index of the optimal arm. Given a new problem, agents who discovered the algorithm of this domain would first try the leaky arm and then go straight to the optimal arm. 
% A similar evaluation setup was introduced in the 3D navigation task presented by \citet{mirowski2016learning}. 
Furthermore, as an acid test, algorithmic generalization also measures the agent's ability in long-term planning in unseen problem configurations, after acquiring adequate information. Evaluation as such has been discussed by \citet{tamar2016value} and \citet{guez2019investigation}. 
% More recently, \citet{xu2019can} study this generalization under the theoretical framework of algorithmic alignment. 

Problem domains discussed above, however, still lack rich concept spaces, nor do they test agents' perceptual generalization, omitting the interaction among the three levels introduced in this paper. Essentially, they are still far-off from the famous Atari game, Frostbite, which is argued to be a testbed for humanlike problem solving \citep{lake2017building}. In this work, we introduce a new problem domain to facilitate joint efforts towards representations with these three levels of generalization.

\setstretch{1}

\begin{figure*}[ht!]
    \centering
    \includegraphics[width=\linewidth]{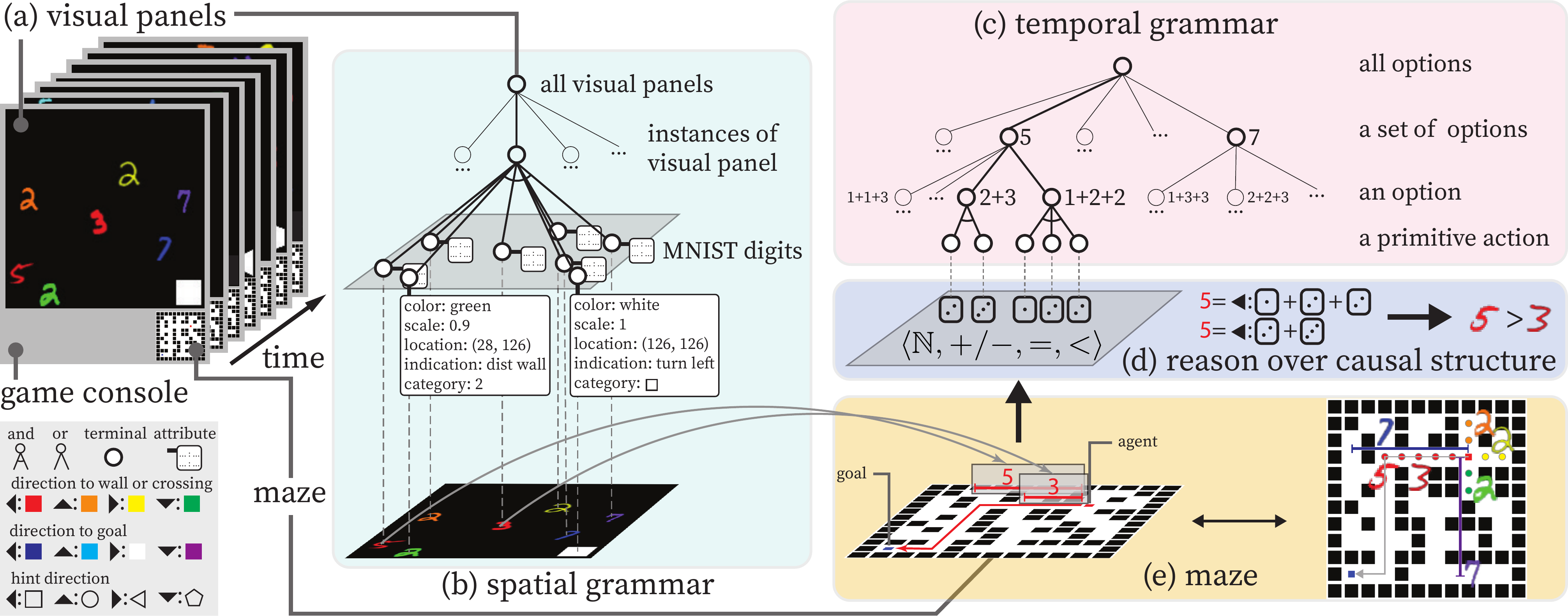}
    \caption{Illustration of the \ac{halma} basics (see \cref{sec:halma_basics}), problem generation, and concept space (see \cref{sec:problem_generation}). (a) Given a visual panel with various colored MNIST digits and a hint, an autonomous agent is tasked to reach the goal in a maze. The concept space guides the generation of the visual panels; it consists of (b) spatial grammar, (c) temporal grammar, and (d) causal structure. (e) The semantics and affordance of the colored MNIST digits are augmented on the corresponding maze; the maze is not shown to the agent.}
    \label{fig:halma_stc}
\end{figure*}

\section{\texorpdfstring{\acf{halma}}{}}

\subsection{\texorpdfstring{\ac{halma}}{} Basics}
\label{sec:halma_basics}

The setup of \ac{halma} is minimal and interpretable. Instead of replicating the entire game of \emph{Halma}, we only preserve the most essential ingredients: The learning agent is cast as one pawn, navigating around the ``magical'' \emph{Halma} landscape by itself. To simplify the environment without lost of generality, we build a maze in a grid-world for each \emph{scenario} (or \emph{problem} henceforth), resembling a \emph{cognitive map} of the agent. Distinct from vanilla grid-world maze games, \ac{halma} is novel in terms of our design of its observation space and action space. The agent perceives neither the global map nor any local patch of the global map; instead, it is shown with a visual panel of various numbers of MNIST digits in various color, randomly scaled and placed; see \cref{fig:halma_stc} (a). These colored digits indicate the \emph{semantics} of (i) the distance till a wall towards each direction, (ii) the distance till the nearest crossing or T-junction towards each direction, and (iii) the distance and direction to the goal; the visual panel only displays non-zero distances. For example, in \cref{fig:halma_stc} (a) (e), $\mnist{5}{0}$ indicates the wall to the left is 5-grid away, and $\mnist{3}{0}$ indicates the nearest crossing is is 3-grid away to the left; the visual color of {\color{red}{red}} refers to the semantics of ``left.'' The agent will also be hinted with a symbol from the set $\{\Circle,\Triangle,\Square,\Pentagon\}$ at any crossing for the correct direction; see an example of $\Square$ in \cref{fig:halma_stc} (a). When making a decision, the agent needs to first select a direction and then select either a primitive action or an option composed by a sequence of primitive actions \citep{sutton1999between} with maximum length $\texttt{max\_opt\_len}$. The direction set is $\{\caretu,\caretd,\caretl,\caretr\}$. The primitive action set, in terms of the number of units to move, is $\{\dicez,\diceone,\dicetwo,\dicethree\}$; this design of primitive numbers with a maximum of three aligns with the doctrine of core knowledge in developmental psychology \citep{feigenson2003tracking,dehaene2011number}. If an option is selected, consecutive hops as in \emph{Halma} are simulated; all observations from intermediate states will be skipped, and only the observation of the final state is provided. A move would fail if a wall stops the agent, leaving the agent's position unchanged; failure moves bring penalties to the agent. The agent would receive a positive reward when reaching the goal. Such a design encourages the agent to comprehend which MNIST digit \emph{affords} it to take which moves.

Essentially, \ac{halma} is a 2D contextual navigation game, sharing the same spirit with those in \citet{mirowski2016learning} and \citet{ritter2018been}. However, \emph{contexts} in these prior works are elusive and conceptually meaningless. As such, they only evaluate generalization at either the visuomotor or algorithmic level. In stark contrast, \ac{halma} is unique, possessing a rich, crisp, and challenging configuration space of problems, semantics, and affordance; see details in the next subsection. 

\subsection{Problem Generation and Concept Space}
\label{sec:problem_generation}

Generating a \ac{halma} problem consists of two sub-procedures: (i) generating a grid-world maze problem with \emph{valid} optimal paths, and (ii) producing a set of visual panels, based on an explicit spatial grammar of the \emph{concept space}, that uniquely represent observations in the maze.

\setstretch{0.95}

Generating a grid-world maze problem is intricate since \ac{halma} is a partially observable game. A randomly generated maze may perplex the agent with ambiguous observations that hinders the agent's formation of a coherent strategy; see \cref{appx:halma_problems} for an example. To alleviate this issue, instead of first generating a complete maze and then producing optimal paths, our solution is to reverse this process by first generating \emph{valid} optimal paths and then adding deceptive branches to construct a grid-world maze. Formally, a path is said to be \emph{invalid} if an agent who possesses an oracle understanding of the concept space fails to make the oracle decision; such a definition of \emph{validity} is deeply rooted in the concept space that the agent is required to learn. We refer the readers to check \cref{appx:halma_problems} for an example of \emph{invalid} optimal path, an example of a successfully generated maze with a \emph{valid} optimal path, an example sequence, and additional implementation details.

Producing visual panels heavily relies on the concept space. The concept space of \ac{halma} consists of an explicit \emph{spatial grammar} for visual panels, an implicit \emph{temporal grammar} for actions and options, and an underlying \emph{causal structure} that specifies the intersection of spatial and temporal grammar. For simplicity, we only introduce them verbally here; see an illustration in \cref{fig:halma_stc} and their formal definitions in \cref{appx:concept_space}. Intuitively, the spatial grammar produces all possible descriptions of visual panels, spanning all configurations of \emph{semantics} introduced in \cref{sec:halma_basics}. To generate a visual panel for a given state, we first sample an MNIST digit for each entry of its description and then sample a random scale and position. The sampled MNIST digit is then colored on the basis of its semantics, \ie, directions to a wall, a crossing, or a goal; see \cref{fig:halma_stc} (b) and the legend. The temporal grammar produces all possible moves, either a single primitive action or a composed option, regardless of the visual stimuli. For instance, a non-terminal node $\caretl:\texttt{5}$ can be parsed into options $\texttt{opt}$, such as $\caretl:\dicetwo+\diceone+\dicetwo$ and $\caretl:\dicetwo+\dicethree$; see \cref{fig:halma_stc} (c). Despite of their distinction in terms of how an option is decomposed into primitive actions, these options are equivalent in their causal effects. Specifically, these causal effects bind visual MNIST digits with digital actions based on one of the simplest mathematical structures in human cognition \citep{flavell1963developmental}: $\langle \mathbb{N}, +/-, =, < \rangle$; namely, \emph{natural numbers} $\mathbb{N}$, \emph{operations} $+/-$, and \emph{relations} $=$, $<$ over $\mathbb{N}$. For example (see also \cref{fig:halma_stc} (d)), a learning agent is expected to understand relations between $\mnist{5}{0}$ and $\mnist{3}{0}$ via 
\begin{itemize}[leftmargin=*,noitemsep,nolistsep]
    \item $\langle\texttt{S},<\rangle$: the set of semantic \emph{generators}\footnote{For the sake of formalism, we adopt the terminology from General Pattern Theory \citep{grenander1993general}, wherein the term \emph{generator} refers to basic units in a \emph{configuration space}. Intuitively, an object file \citep{kahneman1992reviewing}, is a semantic generator. It is also a generator for configuration spaces of affordance and causality, for which actions/options are also generators. We refer the readers to \cref{appx:concept_space} for detailed formal definitions.} with an \emph{order} over it, \eg, $\mnist{3}{0} < \mnist{5}{0}$;
    \item $\langle\texttt{A},{+/-},{=}\rangle$: the set of affordance generators with operations and equality, \eg, $\mnist{5}{0} = \caretl:\dicetwo+\diceone+\dicetwo=\caretl:\dicetwo+\dicethree=\ldots$;
    \item $\langle\texttt{A},{+/-},{<}\rangle$: the set of affordance generators with operations and inequality, \eg, $\caretl:\diceone+\dicetwo<\mnist{5}{0}$, $\mnist{5}{0}<\caretl:\dicetwo+\dicetwo+\dicethree$;
    \item $\langle\texttt{C},{+/-},{=}\rangle$: the set of causal generators with operations and equality, \eg, $\mnist{5}{0}=\mnist{3}{0}+\caretl:\dicetwo$.
\end{itemize}

% One may challenge the deterministic problem setup. We want to argue that environments with deterministic state transitions and rewards are common in daily experience. For example, in navigation, when you exit a room and then return back, you usually end up in the room where you started. 

\subsection{Task Formulation and Evaluation}
\label{sec:task_formulation}

We expect agents who developed the concept space to leverage this knowledge and rapidly solve new problems in \ac{halma}. To this end, we formulate this rapid problem-solving task with an objective to \emph{maximize the agent's rewards accumulated over a few trials in a \textbf{novel} problem instance}: 
\begin{equation}
    \mathbb{E}_\zeta[\sum\nolimits_{i=0}^{N} \gamma^{\sum_{j=0}^{i-1} \text{len}(\tau_{j})} \sum\nolimits_{t=0}^{\text{len}(\tau_i)-1}\gamma^{t} R({s_{\tau_i, t}, a_{\tau_i, t}})].
    \label{eq:task_obj}
\end{equation}
Specifically, an agent's experience in each problem instance is dubbed an \emph{episode} $\zeta$ \citep{wang2016learning}, which terminates when a maximum number of \emph{steps} $L$ is reached or a maximum number of \emph{trials} $N$ have been accomplished. A \emph{trial} $\tau$ proceeds with actions $a_{\tau, t}$, spanning multiple \emph{steps} $t$; it starts from an initial state $s_0$ and terminates when the agent reaches the goal $s_g$ (thus accomplished), or when it consumes the maximum number of steps $H$ (thus failed). The agent is respawned to the initial state when a trial terminates. It is awarded $R(s_g, \cdot)$ if the trial is accomplished. The cumulative reward in one episode is the sum of temporally $\gamma$ decayed accomplishments. When one episode terminates, the agent is presented with the next problem. 

Under this task formulation, learning agents should be evaluated against oracle solutions, analogous to ground-truth annotations in supervised learning; recall that the oracle agent has complete understanding of the concept space and the problem domain. Since \ac{halma} is a \emph{partially observable} domain, its oracle behavior consists of two aspects: optimal exploration and optimal planning. As introduced in \cref{sec:problem_generation}, problems are generated by adding deceptive branches to optimal paths. Hence, the optimal exploration strategy is to stop at each crossing to obtain the hint from the visual panel. Intuitively, the agent should understand ``when two digits with the same color are exhibited in the visual panel, the \emph{lesser} one indicates the crossing, and I should stop there for hint'' 
% This is an abstract version of crossing detection in visual 3D navigation, which requires the agent to understand 
based on the concept of $\langle\texttt{S},<\rangle\cup\langle\texttt{A},{+/-},{<}\rangle$. An oracle agent would sacrifice the first trial to explore; note that the cost is still low as it would explore along the optimal path with the guidance of hints, avoiding all deceptive branches. Afterwards, the oracle agent should retrieve its experience and merges consecutive moves towards the same direction to form the optimal plan. Take the maze example shown in \cref{fig:halma_stc} (e); during exploration, the agent sees a $\mnist{5}{0}$ and a $\mnist{3}{0}$ in the visual panel and takes an option $\caretl:\diceone+\dicetwo$ to obtain a hint $\Square$, which guides it to keep moving left $\caretl:\dicetwo$ until the wall. Then in the second trial, the agent should exploit $\langle\texttt{A},{+/-},{=}\rangle\cup\langle\texttt{C},{+/-},{=}\rangle$ via $\caretl:\diceone+\dicetwo+\dicetwo$. With this oracle agent, we can have evaluation metrics normalized across different problems. Instead of directly calculating the ratio of \cref{eq:task_obj} between proposed agents and the oracle agent, which involves strong non-linearity, we carefully decompose it into three metrics with more intuitive measures:
\begin{itemize}[leftmargin=*,noitemsep,nolistsep]
    \item Ratio of valid moves $\bm{\rho_a}=\mathbb{E}_\zeta[\frac{\#\text{valid moves}}{\sum_i\text{len}(\tau_i)}]$ for semantics and affordance understanding;
    \item Success rate of goal reaching $\bm{\rho_g}=\mathbb{E}_\zeta[\frac{1}{N}\sum_i \delta(s_{\tau_i, -1}=s_g)]$ for leveraging concepts to explore;
    \item Efficiency in exploration and planning $\bm{\rho_p}=\mathbb{E}_\zeta[\frac{1}{N}\sum_i \frac{\text{len}(\tau^{\star})}{\text{len}(\tau_i)}]$ for algorithmic understanding. 
\end{itemize}
% \item Inference time on novel problem instances

\subsection{Generalization Test}
\label{sec:gen_test}

One of our key contributions in \ac{halma} is a novel paradigm to test agents' capability in all three levels of generalization, which extends the classical paradigm of statistical learning. Our training set consists of 100 mazes\footnote{This design reflects our thesis argument, \ie, agents shall generalize their understanding from limited exposure to the concept space. An ablation study on the volume of training set can be found in \cref{appx:ablation_trainset}.} along with their visual panels; we summarize the statistics of these visual panels in \cref{appx:stats} to show that the generated dataset is balanced, yielding fair distributions of crucial statistics. Different from the classic paradigm, the evaluation of agent's performance in \ac{halma} would emphasize on the \emph{explicit extrapolation} test, which should be conducted in the \emph{held-out} compositional and relational configurations; such design echoes recent trend in evaluating agent's generalization capability \citep{burgess2019monet,lake2018generalization,zambaldi2018deep}. Compared to these prior domains, \ac{halma} is unique as it is a partially observable and interactive problem-solving task, wherein an agent is tasked to \emph{autonomously} learn the immense concept space and form the abstract knowledge. Hence, simply holding off a \emph{pre-selected}, \emph{fixed} subset of conceptual configurations would impose severe restrictions on problem generators. For instance, if we would like to allow agents to see a $\mnist{4}{0}$, they must be able to see a $\mnist{3}{0}$ by simply moving $\caretl:\diceone$ from where they see $\mnist{4}{0}$. In other words, if we managed to strictly withhold $\mnist{3}{0}$ from agents, they would not see any red digits larger than 3 in this \emph{interactive} problem solving task. Therefore, an \emph{ex post} evaluation protocol that \emph{dynamically} generates tests is more desirable.
% Due to the combinatorial nature of the problem space and the concept space, holding a set of conceptual configurations would impose severe restrictions on problem generators, incurring logistical issues. 

In this paper, we propose an ingenious solution: Instead of \emph{aimlessly} generating a large test set of \emph{random} cases, we devise an algorithm to \emph{proactively} generate \emph{tailored} tests in accord to what the agent might have learned; this design would produce a definitive and much more informative evaluation of agent's competence. The intuition is simple: When a teacher finds a student consistently make right decisions during training, wherein the student only needs to understand $\mnist{3}{1} < \mnist{5}{1}$ and $\mnist{4}{0}=\mnist{2}{0}+\caretl:\dicetwo$, the teacher may quiz the student on $\mnist{3}{0}$ vs $\mnist{5}{0}$ and $\mnist{2}{1}$ vs $\mnist{4}{1}$. To implement this protocol in \ac{halma}, we first store agents' experience during training as their external memory $\texttt{MEM}$. We then construct a representation to emulate agents' \emph{knowledge bases} (KB) for $\langle\texttt{S},<\rangle$ and $\langle\texttt{A},{+/-},{=}\rangle\cup\langle\texttt{C},{+/-},{=}\rangle$: $\texttt{KB}_{\texttt{S}}$ tracks the agent's understood configurations on semantics, and $\texttt{KB}_{\texttt{A}\lor\texttt{C}}$ tracks the agent's understood configurations on affordance and causality. Here, we assume that (i) valid decisions\footnote{Note that some decisions may come from random exploration. We introduce a threshold on the visitation count to filter them out.} in experience were made upon understanding \emph{inequality} configurations, and (ii) agents understand configurations involving \emph{equality} and \emph{operations} in experienced transitions. With these KBs, we \emph{dynamically} generate test problems with novel configurations, wherein agents should likewise act appropriately if they understood not only seen configurations but their underlying concepts; see details of constructing KBs and generating test problems in \cref{appx:kb}.

Tests in \ac{halma} are on the competence basis: Conceptual generalization is built upon perceptual generalization, with the algorithmic generalization resides on top. Tests for perceptual generalization are backed by the \emph{spatial grammar}, including unseen MNIST images and unseen compositions of visual attributes, \ie, shape and color. Tests for conceptual generalization are based on \emph{the concept of $\langle \mathbb{N}, +/-, _, =, < \rangle$}, consisting of novel \emph{equality} and \emph{inequality} configurations. Results of these two tests are manifested in algorithmic generalization. Specifically, agents could only pass all of these tests by making right \emph{exploration} decisions based on relations of novel digit pairs $\langle \texttt{d}_\texttt{1}, \texttt{d}_\texttt{2} | \texttt{type}\rangle$, where \texttt{type} refers to various directions. Inappropriate exploration may cause agent to miss hints at crossings or to be trapped in dead-ends, resulting in failures of the tests. Moreover, these novel digit pairs also test the agents' understanding of the \emph{temporal grammar}, requiring agents to make proper \emph{exploitation} decisions by merging novel consecutive actions/options into a \emph{greater} option.

Since conceptual generalization connects the other two, all three levels of generalization are covered when test problems are dynamically generated with novel configurations in $\langle \mathbb{N}, +/-, =, < \rangle$. Recall that the generation mechanism of a problem is to first generate an unseen configuration of optimal path and then add deceptive branches; the latter is pivotal for a test problem since it involves generating novel digit pairs $\langle \texttt{d}_\texttt{1}, \texttt{d}_\texttt{2} | \texttt{type}\rangle$. 
% whose inequality relation was not required during training but should be understood by the agent. 
By design, the lesser digit within a pair should indicate the distance to the nearest crossing, and the greater the distance to the wall. Hence, agents could be tested by these novel digit pairs, queried based on the agent's KBs. We categorize the problems into:
\begin{itemize}[leftmargin=*,noitemsep,nolistsep]
    \item \underline{S}emantic \underline{T}est (\textbf{ST}): $\texttt{KB}_{\texttt{ST}}=(\langle \texttt{d}_\texttt{1}, \texttt{d}_\texttt{2}|\texttt{type}\rangle\not\in \texttt{KB}_{\texttt{S}})\land(\exists_\texttt{x}\langle \texttt{d}_\texttt{1}, \texttt{d}_\texttt{2}|\texttt{x}\rangle\in \texttt{KB}_{\texttt{S}})$, \ie, testing visual panels differentiated from $\texttt{KB}_{\texttt{S}}$ in terms of color, shape, or other MNIST digits.
    \item \underline{Af}fordance \underline{T}est (\textbf{AfT}): $\texttt{KB}_{\texttt{AfT}}=(\forall_{\texttt{x}}\langle \texttt{d}_\texttt{1}, \texttt{d}_\texttt{2}|\texttt{x}\rangle\not\in \texttt{KB}_{\texttt{S}})\land((\exists\langle \texttt{d}_\texttt{1}, \texttt{d}_\texttt{2}|\texttt{x}\rangle\in \texttt{KB}_{\texttt{A}\lor\texttt{C}})\lor(\texttt{d}_\texttt{1}=\texttt{opt}_\texttt{1}\in \texttt{KB}_{\texttt{A}\lor\texttt{C}}\land \texttt{d}_\texttt{2}=\texttt{opt}_\texttt{2}\in \texttt{KB}_{\texttt{A}\lor\texttt{C}}))$, \ie, testing inequalities inferred from equalities in $\texttt{KB}_{\texttt{A}\lor\texttt{C}}$. $\texttt{opt}$ denotes actions or options. 
    \item \underline{An}alogy \underline{T}est (\textbf{AnT}): $\texttt{KB}_{\texttt{AnT}}=(\forall_{\texttt{x}}\langle \texttt{d}_\texttt{1}, \texttt{d}_\texttt{3}|\texttt{x}\rangle\not\in \texttt{KB}_{\texttt{ST}\lor\texttt{AfT}})\land(\exists\{\langle \texttt{d}_\texttt{1}, \texttt{d}_\texttt{2}|\texttt{x}\rangle,\langle \texttt{d}_\texttt{2}, \texttt{d}_\texttt{3}|\texttt{x}\rangle\}\subset \texttt{KB}_{\texttt{ST}\lor\texttt{AfT}})\land(\exists\{\langle \texttt{d}_\texttt{1}', \texttt{d}_\texttt{2}'|\texttt{x}\rangle, \langle \texttt{d}_\texttt{2}', \texttt{d}_\texttt{3}'|\texttt{x}\rangle, \langle \texttt{d}_\texttt{1}', \texttt{d}_\texttt{3}'|\texttt{x}\rangle\}\subset \texttt{KB}_{\texttt{ST}\lor\texttt{AfT}})$, \ie, testing inequalities inferred from the \emph{transitivity} of $<$. $\texttt{KB}_{\texttt{ST}\lor\texttt{AfT}}=\texttt{KB}_{\texttt{ST}}\cup\texttt{KB}_{\texttt{AfT}}$.
    % \item \underline{R}ecursive \underline{An}alogy \underline{T}est (\textbf{RAnT}): inequalities can only be inferred with mathematical induction.
\end{itemize}
Specific examples of these tests can be found in \cref{tab:exp_results}. See \cref{appx:kb} for detailed explanation. 

\section{Models and Experiments}

The motivating questions of our experiments are: (i) Do model-free agents, exploiting generic inductive biases, develop concepts that generalize in a way, akin to human knowledge? (ii) If there are indeed certain meta-benefits induced by these architectural priors towards problem solving, are they achievable with only limited exposure to the concept space? As it is logistically challenging to experiment with all existing models, a representative subset is culled for benchmark: model-free reinforcement learning agents \citep{wang2016learning,zambaldi2018deep} with gated memory mechanism \citep{hochreiter1997long}, self-attention mechanism \citep{vaswani2017attention}, or both. Notably, \citet{wang2016learning} argued that when an RNN agent is fed with previous actions and rewards, its LSTM module would emulate an inner reinforcement learning algorithm; the agent is thus learning to reinforcement learn. They demonstrated that the learned exploration strategy is more efficient than a near-optimal model-free exploration algorithm. 
% Similar arguments on LSTM agents learning to plan given full observation were made by \citet{guez2019investigation}. 
\citet{zambaldi2018deep} argued that by exploiting stacked attention modules, Transformer agents can conduct iterated reasoning with seen relational units and generalize to unseen scenarios. 
By our evaluation protocol, however, these prior models did not demonstrate conclusive evidence to support all three levels of generalization proposed in this paper; hence, the precise level of generalization is obscure. Crucially, neither of them evaluated the learned agents \emph{under limited exposure} to a \emph{complex concept space} as in \ac{halma}.

\cref{tab:exp_results} shows the full list of agents used in our experiments; see \cref{appx:models} for implementation details. All agents are trained with an off-the-shelf reinforcement learning method, TD3 \citep{fujimoto2018addressing}. All agents' policies converged at the end of training.

To decouple the evaluation of conceptual generalization from perceptual generalization, we first conduct experiments with symbolic one-hot observations, which can be regarded as the ground-truth representation of perception; see details of this observation space in \cref{appx:exp_protocol}. All agents show relatively low valid move ratio $\bm{\rho_a}$ in tests of random split, indicating their understanding of affordance is brittle even with the ground-truth semantics. Under this precondition, we find that all agents can still perform relatively well in terms of goal-reaching $\bm{\rho_g}$ and efficiency $\bm{\rho_p}$ in random splits. However, when transferred to our generalization tests, MLP agents exhibits a significant degradation. Agents with LSTM modules, on the contrary, can somehow maintain or even surpass their $\bm{\rho_g}$ and $\bm{\rho_p}$ in training problems. One possible explanation to their high $\bm{\rho}_g$ is: With a memory mechanism, they learn to recover from dead-ends even if they missed the hints at crossings. Even though they also have higher $\bm{\rho}_p$ than MLP agents, consistent with the findings reported by \citet{wang2016learning}, this measure is still disconcertingly low. Such low performance implies that agents do not understand the concept space well, especially in terms of the temporal grammar. Transformer agents do perform better than MLP agents in generalization tests, but not as good as LSTM agents. In particular, even though \citet{zambaldi2018deep} argued that Transformer agents as such may learn to plan, their lower $\bm{\rho_p}$ in \ac{halma} task implies the opposite, at least under partial observation without a memory mechanism. Combining the benefits from the attention and the memory mechanisms, TRAN+LSTM agents outperform others in almost all generalization tests on both $\bm{\rho_g}$ and $\bm{\rho_p}$. Another interesting phenomenon is: By removing the constraint of \emph{limited exposure} (\eg, we increase the training volume to $10\times$), all agents, no matter what inductive biases are encoded, achieve around $80\%$ measured by $\bm{\rho_g}$, and those with LSTM modules have $\bm{\rho_p}$ at around $45\%$; see details in \cref{appx:ablation_trainset}. Since no state-of-the-art agents could pass the test on $\bm{\rho_p}$, we summarize the results of symbolic experiments as: In the spectrum of model-based vs model-free, emerged strategies still reside on the model-free side of the oracle agent. Significant efforts are needed to devise agents capable of humanlike conceptual and algorithmic generalization. 

Under visual observation, however, all agents fail the generalization test when simply connected with a convolutional module, even in the easiest setup (\texttt{max\_opt\_len=1}). Assuming CNNs do not offer sufficient priors to induce an object-oriented, independently disentangled representation, we pretrain a state-of-the-art multi-object segmentation and disentanglement model, SPACE \citep{lin2019space}, with all visual panels in the training set. The converged model exhibits remarkable generalization in reconstruction, segmentation, and detection, consistent with the results reported by \citet{lin2019space}; see details in \cref{appx:space}. One would expect that, by connecting the encoder of this powerful pretrained visual module with an RL agent using a Transformer module for the object-oriented encoding, the model would have a superb performance. Counter-intuitively, our results show that SPACE agents perform worse than CNN+TRAN agents even under random split. A further investigation reveals that the latent space of object slots fails to disentangle shapes or colors (\eg, $\mnist{3}{5}$ vs $\mnist{5}{6}$), even though they can be substantially distinguished and reconstructed by the strongly nonlinear decoder. This explanation also accounts for SPACE agents' low valid move ratio in test problems ($\bm{\rho_a}=41.62\pm1.20$). In principle, they misunderstand affordance because they fail to recognize ``what it is'' in the first place. More details on this SPACE experiment can be found in \cref{appx:space}. Taking together, we argue that \ac{halma} does extend the evaluation paradigm of perceptual generalization, posing new challenges to the community of unsupervised disentanglement. 

\begin{table}[t!]
    \caption{Examples and results of generalization tests (- indicates no problem is dynamically generated)}
    \label{tab:exp_results}
    \resizebox{\linewidth}{!}{
        \begin{tabular}{c@{\hskip9pt}cccccc|ccc}
        \toprule
        \multicolumn{2}{c}{\multirow{3}{*}{Test Type \& Examples}} & \multirow{3}{*}{} & \multicolumn{7}{c}{Models \& Results} \\ \cline{4-10} 
        \multicolumn{2}{c}{} & & \multicolumn{4}{c}{SYMBOLIC \small{(\texttt{max\_opt\_len=5})}} & \multicolumn{3}{c}{VISUAL \small{(\texttt{max\_opt\_len=1})}} \\ 
        \multicolumn{2}{c}{} &\% & MLP & \multicolumn{1}{c}{LSTM} & \multicolumn{1}{c}{TRAN} & \multicolumn{1}{c}{TRAN+LSTM} & CNN+MLP & \multicolumn{1}{c}{CNN+TRAN} & \multicolumn{1}{c}{SPACE} \\ \hline
        \parbox[t]{2mm}{\multirow{3}{*}{\rotatebox[origin=c]{90}{T}}} & \multirow{3}{*}{Training problems}& $\bm{\rho_a}\uparrow$ & \ditem{94.78}{4.11} & \ditem{87.88}{2.14} & \ditem{85.43}{6.77} & \ditem{86.95}{3.09} & \ditem{85.61}{7.22} & \ditem{89.71}{2.61} & \ditem{83.55}{2.65}\\
        & & $\bm{\rho_g}\uparrow$ & \ditem{99.23}{0.63} & \ditem{57.22}{3.07} & \ditem{93.85}{1.26} & \ditem{72.33}{5.79} & \ditem{75.76}{4.77} & \ditem{58.33}{4.19} & \ditem{16.33}{0.94}\\
        & & $\bm{\rho_p}\uparrow$ & \ditem{71.67}{1.73} & \ditem{50.91}{3.54} & \ditem{67.89}{0.63} & \ditem{63.97}{5.84} & \ditem{63.77}{2.68} & \ditem{35.31}{3.00} & \ditem{12.02}{1.17}\\ \hline
        \parbox[t]{2mm}{\multirow{3}{*}{\rotatebox[origin=c]{90}{RT}}} & \multirow{3}{*}{Random split} & $\bm{\rho_a}\uparrow$ & \ditem{62.98}{1.52} & \ditem{76.09}{2.10} & \ditem{65.15}{4.45} & \ditem{62.31}{2.90} & \ditem{13.30}{2.30} & \ditem{43.09}{7.92} & \ditem{41.62}{1.20}\\
        & & $\bm{\rho_g}\uparrow$ & \ditem{51.00}{2.21} & \ditem{57.78}{3.49} & \ditem{82.82}{0.96} & \ditem{54.00}{2.94} & \ditem{7.58}{0.43} & \ditem{14.00}{4.24} & \ditem{3.67}{0.47}\\
        & & $\bm{\rho_p}\uparrow$ & \ditem{54.91}{2.85} & \ditem{45.15}{1.46} & \ditem{58.07}{1.01} & \ditem{40.13}{2.52} & \ditem{5.09}{1.17} & \ditem{8.33}{1.96} & \ditem{2.66}{0.19}\\ \hline
        \parbox[t]{2mm}{\multirow{4}{*}{\rotatebox[origin=c]{90}{ST}}} & $\langle\mnist{3}{1}, \mnist{5}{1}, \mnist{1}{0}, \mnist{7}{5}\rangle\in\texttt{MEM}$, & $\bm{\rho_g}\uparrow$ & \ditem{55.00}{7.07} & \ditem{50.00}{8.16} & \ditem{41.67}{8.50} & \ditem{66.67}{13.12} & \ditem{0.00}{0.00} & \ditem{0.00}{0.00} & \ditem{0.00}{0.00}\\
        & test $\langle\mnist{3}{1},\mnist{5}{1}, \mnist{4}{3}, \mnist{2}{6}\rangle\not\in\texttt{MEM}$. & $\bm{\rho_p}\uparrow$ & \ditem{19.90}{2.18} & \ditem{24.02}{7.20} & \ditem{16.34}{3.90} & \ditem{35.74}{5.85} & \ditem{0.00}{0.00} & \ditem{0.00}{0.00} & \ditem{0.00}{0.00}\\\cline{2-10}
        & $\mnist{3}{1} < \mnist{5}{1}\in\texttt{KB}_\texttt{S}$, & $\bm{\rho_g}\uparrow$ & \ditem{25.00}{8.16} & \ditem{63.33}{6.24} & \ditem{43.33}{6.23} & \ditem{78.33}{2.36} & \ditem{0.00}{0.00} & \ditem{0.00}{0.00} & \ditem{0.00}{0.00}\\
        & test $\langle\mnist{3}{3},\mnist{5}{3}\rangle\not\in\texttt{KB}_\texttt{S}$. & $\bm{\rho_p}\uparrow$ & \ditem{7.37}{2.33} & \ditem{26.31}{2.34} & \ditem{12.22}{1.83} & \ditem{34.79}{4.25} & \ditem{0.00}{0.00} & \ditem{0.00}{0.00} & \ditem{0.00}{0.00}\\ \hline
        \parbox[t]{2mm}{\multirow{8}{*}{\rotatebox[origin=c]{90}{AfT}}} & $\mnist{5}{0}=\mnist{3}{0}+\caretl:\dicetwo\in\texttt{KB}_{\texttt{A}\lor\texttt{C}},$ & $\bm{\rho_g}\uparrow$ & \ditem{41.67}{2.36} & \ditem{60.00}{10.80} & \ditem{36.67}{8.50} & \ditem{58.33}{10.27} & \ditem{0.00}{0.00} & \ditem{0.00}{0.00} & \ditem{0.00}{0.00}\\
        & test $\langle\mnist{3}{0},\mnist{5}{0}\rangle\not\in\texttt{KB}_\texttt{S}$. & $\bm{\rho_p}\uparrow$ & \ditem{15.10}{0.35} & \ditem{28.91}{7.62} & \ditem{14.01}{3.75} & \ditem{27.11}{2.12} & \ditem{0.00}{0.00} & \ditem{0.00}{0.00} & \ditem{0.00}{0.00}\\\cline{2-10}
        & $\{\mnist{5}{1} = \caretu:\dicetwo+\dicethree,\mnist{3}{1} = \caretu:\dicethree\}$ & $\bm{\rho_g}\uparrow$ & \ditem{31.67}{8.50} & \ditem{45.00}{10.80} & \ditem{43.33}{6.24} & \ditem{71.67}{6.24} & \ditem{0.00}{0.00} & \ditem{0.00}{0.00} & \ditem{0.00}{0.00}\\
        & $\subset \texttt{KB}_{\texttt{A}\lor\texttt{C}}$, test $\langle\mnist{3}{1},\mnist{5}{1}\rangle\not\in\texttt{KB}_\texttt{S}$. & $\bm{\rho_p}\uparrow$ & \ditem{11.68}{3.34} & \ditem{17.15}{5.82} & \ditem{17.86}{3.02} & \ditem{35.40}{3.71} & \ditem{0.00}{0.00} & \ditem{0.00}{0.00} & \ditem{0.00}{0.00}\\\cline{2-10}
        & $\mnist{5}{1}=\mnist{3}{1}+\caretu:\dicetwo\in\texttt{KB}_{\texttt{A}\lor\texttt{C}},$ & $\bm{\rho_g}\uparrow$ & \ditem{6.67}{2.36} & \ditem{100.00}{0.00} & \ditem{25.00}{0.00} & - & \ditem{0.00}{0.00} & \ditem{0.00}{0.00} & \ditem{0.00}{0.00}\\
        & test $\langle\mnist{3}{0},\mnist{5}{0}\rangle\not\in\texttt{KB}_\texttt{S}$. & $\bm{\rho_p}\uparrow$ & \ditem{1.48}{0.52} & \ditem{51.86}{0.18} & \ditem{5.83}{0.24} & - & \ditem{0.00}{0.00} & \ditem{0.00}{0.00} & \ditem{0.00}{0.00}\\\cline{2-10}
        & $\{\mnist{5}{2} = \caretr:\dicetwo+\dicethree,\mnist{3}{3} = \caretd:\dicethree\}$ & $\bm{\rho_g}\uparrow$ & \ditem{0.00}{0.00} & \ditem{86.67}{9.43} & \ditem{50.00}{0.00} & - & \ditem{0.00}{0.00} & \ditem{0.00}{0.00} & \ditem{0.00}{0.00}\\
        & $\subset \texttt{KB}_{\texttt{A}\lor\texttt{C}}$, test $\langle\mnist{3}{0},\mnist{5}{0}\rangle\not\in\texttt{KB}_\texttt{S}$. & $\bm{\rho_p}\uparrow$ & \ditem{0.00}{0.00} & \ditem{29.89}{2.18} & \ditem{10.00}{0.00} & - & \ditem{0.00}{0.00} & \ditem{0.00}{0.00} & \ditem{0.00}{0.00}\\ \hline
        \parbox[t]{2mm}{\multirow{3}{*}{\rotatebox[origin=c]{90}{AnT}}} & $\{\mnist{3}{0}<\mnist{4}{0}, \mnist{4}{1}<\mnist{5}{1}, \underline{\mnist{3}{2}<\mnist{5}{2}}, \mnist{6}{3}$ & $\bm{\rho_g}\uparrow$ & \ditem{35.00}{7.07} & \ditem{48.33}{4.71} & \ditem{41.67}{2.36} & \ditem{41.67}{13.12} & \ditem{0.00}{0.00} & - & \ditem{0.00}{0.00}\\
        & $<\mnist{7}{3},\mnist{7}{1}<\mnist{8}{1}\}\subset \texttt{KB}_{\texttt{ST}\lor\texttt{AfT}}$, & $\bm{\rho_p}\uparrow$ & \ditem{12.19}{1.84} & \ditem{21.84}{0.53} & \ditem{14.45}{1.66} & \ditem{22.03}{6.64} & \ditem{0.00}{0.00} & - & \ditem{0.00}{0.00}\\
        & test $\langle\mnist{6}{0},\mnist{8}{0}\rangle\not\in\texttt{KB}_{\texttt{ST}\lor\texttt{AfT}}$. & & & & & & & & \\ %\hline
        % \parbox[t]{2mm}{\multirow{4}{*}{\rotatebox[origin=c]{90}{RAnT}}} & $\{\mnist{3}{1}<\mnist{4}{1}, \mnist{4}{2}<\mnist{5}{2}, \mnist{5}{1}<\mnist{6}{1}$ & $\rho_g$ & & & & & & & \\
        % & $\mnist{6}{3}<\mnist{7}{3},\mnist{7}{5}<\mnist{8}{5}, \underline{\mnist{3}{0}<\mnist{5}{0}}$ & & & & & & & & \\
        % & $\underline{\mnist{3}{6}<\mnist{6}{6}}\}\subset \texttt{KB}_{\texttt{ST}\lor\texttt{AfT}}$, test & $\rho_p$ & & & & & & & \\
        % & $\langle\mnist{3}{1},\mnist{8}{1}\rangle\not\in\texttt{KB}_{\texttt{ST}\lor\texttt{AfT}\lor\texttt{AnT}}$. & & & & & & & & \\
        \bottomrule
    \end{tabular}
    }%
\end{table}

\section{Related Work}

Recently, there emerges a burst-out of benchmarks for diagnosing a set of clearly defined competencies of AI systems, which we draw inspiration from and sincerely honor. In a word, \ac{halma} differentiates from all of them in its holistic evaluation towards all three levels of generalization.

Readers may be curious about the relation between \ac{halma} and conventional navigation tasks such as \citet{mirowski2016learning}. We hope we have made it clear the difference between \ac{halma} and them in \cref{sec:halma_basics} of main text: In these navigation tasks, there is only one maze, and new problem instances are simply new combinations of initial and goal states. Hence, rapid problem solving only requires agents to memorize the whole maze, whereas in \ac{halma} the only shared structure between problem instances is the concept space. Going beyond memorization, \ac{halma} requires two extra cognitive abilities---understanding and reasoning. We also notice that in another embodied navigation task, the Habitat challenge \citep{savva2019habitat}, agents are indeed evaluated in completely unseen environments, under the protocol of which \citet{wijmans2019dd} has achieved close-to-optimal performance with large-scale training. However, without a clearly specified concept space, the evaluation in Habitat is akin to the Random Split in HALMA under the setup of \texttt{max\_opt\_len=1}. The reason why we emphasize \texttt{max\_opt\_len} is that the very idea of \emph{affordance} is only interesting if the action/option space is large enough and highly structured. Otherwise, when \texttt{max\_opt\_len=1}, agents with memory or attention do generalize well in both Random Split and our Dynamic Test; see detailed results in \cref{appx:ablation_alen}. Perhaps the notion of \emph{affordance} seems a bit abstract in \ac{halma} and can be more intuitive in visual semantic navigation and control \citep{yang2018visual,chaplot2019embodied}. We hope our work can inspire the future development of benchmarks for these topics. 

\ac{clevr} \citep{johnson2017clevr} is one of the earliest datasets that diagnose models' visual reasoning abilities. High-level reasoning skills required in \ac{clevr} include counting, comparing, logical inference, and memory. The same set of skills are also required in \ac{halma}, but without the guidance of language. Accounting for a similar purpose, \citet{bahdanau2018systematic} propose a minimalist alternative, \ac{sqoop}. While relations in \ac{sqoop} are only spatial, benchmarks inspired by \ac{raven} are proposed towards abstract visual reasoning \citep{barrett2018measuring,zhang2019raven}, in which the capacity of sequential decision making is not required. In sum, all prior works listed in this paragraph are discriminative tasks. Different from them, the generative nature of interactive problem solving in \ac{halma} is akin to human exploration in the open-ended world. 

As for planning and reinforcement learning, Box-World and StarCraft II minigames \citep{vinyals2017starcraft} in \citet{zambaldi2018deep} are tasks that also require relational concept learning; the concepts within, however, are mostly spatial.
% , highly dependent on the coordinates of each patch. 
% , which are given as auxiliary input. 
In contrast, the concept space in \ac{halma} is abstract and complex. The mapping from the visual space to the semantic space is non-trivial to learn, which requires agents' understanding of the temporal grammar and the causal structure. Moreover, 
% generalization tests in \citet{zambaldi2018deep} only involve concepts that are not \emph{required} for optimal plans. It is very possible that the agent has \emph{seen} this concept during its exploration. In \ac{halma}, we make the extrapolation test explicit, generating test cases according to agents' experience. In addition, 
\ac{halma} is a partially observable domain that requires dedicated efforts for exploration. 

The closest one that is also inherently generative, compositional, and abstract is the \ac{scan} \citep{lake2018generalization}, an instruction following task. Essentially, \ac{scan} is seq2seq translation, with little uncertainty or variation in primitives. Hence, it does not test agents' perceptual generalization or algorithmic generalziation. In contrast, \ac{halma} is a task for visual concept development and rapid problem solving. Agents need to understand concepts from visuomotor experience and make smart decisions to acquire utility. 

% Ones may consider one recent benchmark, \ac{atari} \citep{anand2019unsupervised}, to be somehow related to \ac{halma}. Though Anand and colleagues also expose the internal states as us, their benchmark aims at unsupervised perceptual disentanglement, thus orthogonal to our work. But we believe their evaluation metrics for semantic disentanglement, alongside with those summarized by \citet{locatello2019challenging}, do get colleagues from our community more equipped in probing perceptual models in all visual reasoning benchmarks, including \ac{halma}.

\section{General Discussions}

In spite of its synthetic nature, we believe \ac{halma} is an impeccable testbed for rapid problem solving that resembles real-world ones. The dedicated design of its internal state facilitates in-depth and comprehensive analyses on agents' capacity in concept development, abstract reasoning, and meta learning that are otherwise impossible with existing problem-solving tasks. Agents can only pass the dynamically generated generalization tests if they possess adequate capacity to \emph{understand} the abstract structure of this task and build a powerful solver upon this understanding. Our experiments demonstrate the inefficacy of model-free reinforcement learning agents in generalizing their understanding, even when incorporated with generic inductive biases. Towards this end, we would like to invite colleagues across the machine learning community to join our challenge.

% It is, on the contrary, much harder to design real-world tasks to evaluate systematic generalization in either conceptual or algorithmic level. But the breakthrough in performance on those tasks could be more impactful. Therefore, there should be a synergy between the study on synthetic tasks and real-world tasks. \ac{halma} should be studied in conjunction with other problem-solving tasks and reinforcement learning tasks towards an agent for general problem-solving.


\subsubsection*{Acknowledgments}
The authors thank Chi Zhang and Baoxiong Jia of UCLA Computer Science Department for useful discussions. The work reported herein was supported by ONR MURI grant N00014-16-1-2007, ONR N00014-19-1-2153, and DARPA XAI N66001-17-2-4029.

\clearpage
\setstretch{1}
\bibliographystyle{iclr2021_conference}
\bibliography{reference}

\clearpage
\resumetocwriting
\renewcommand\thefigure{S\arabic{figure}}
\setcounter{figure}{0}
\renewcommand\thetable{S\arabic{table}}
\setcounter{table}{0}
\pagenumbering{arabic}% resets `page` counter to 1
\renewcommand*{\thepage}{S\arabic{page}}
\setstretch{1}
\appendix
\tableofcontents

\clearpage
\section{Problem Space of \texorpdfstring{\ac{halma}}{}}
\label{appx:halma_problems}

\subsection{Validity of Optimal Paths}

\begin{figure}[th!]
    \centering
    \begin{subfigure}[b]{0.26\linewidth}
        \centering
        \includegraphics[width=0.9\linewidth]{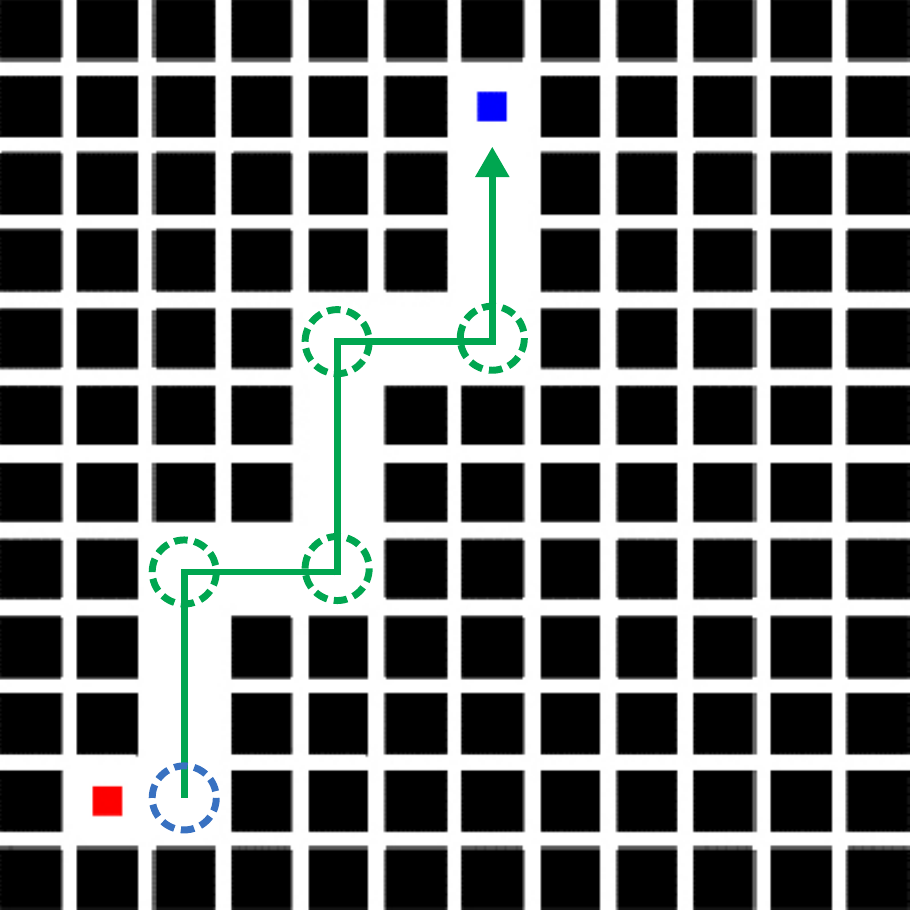}
        \caption{}
    \end{subfigure}%
    \begin{subfigure}[b]{0.26\linewidth}
        \centering
        \includegraphics[width=0.9\linewidth]{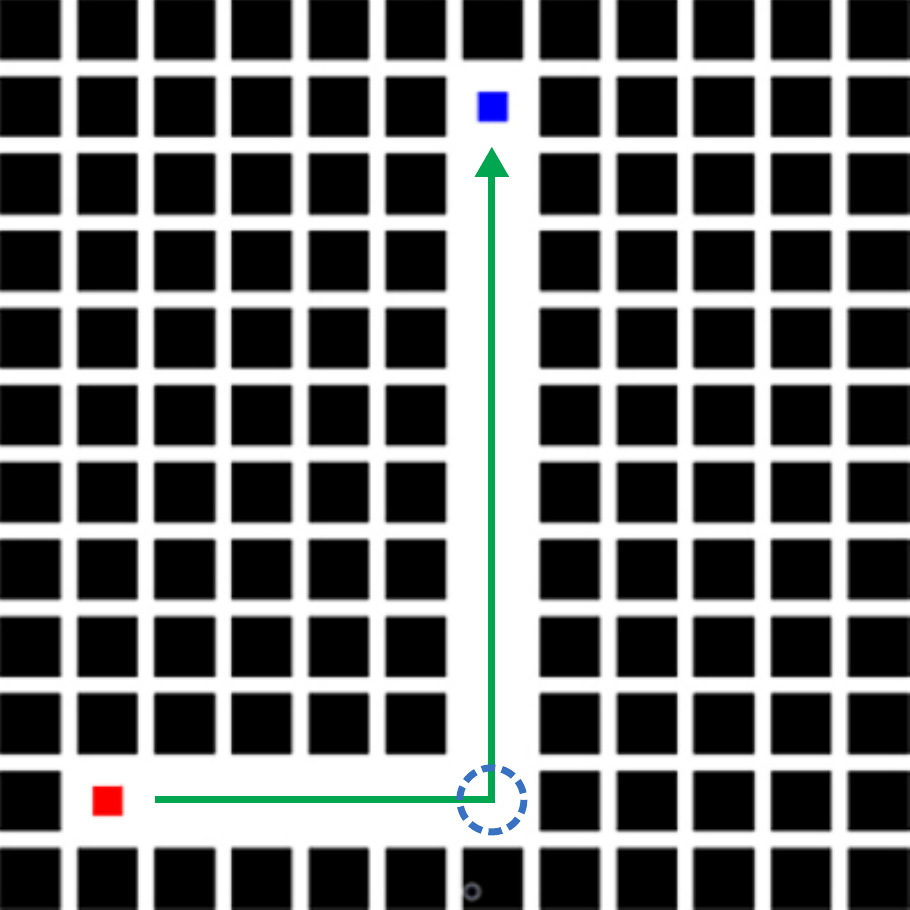}
        \caption{}
    \end{subfigure}%
    \begin{subfigure}[b]{0.26\linewidth}
        \centering
        \includegraphics[width=0.9\linewidth]{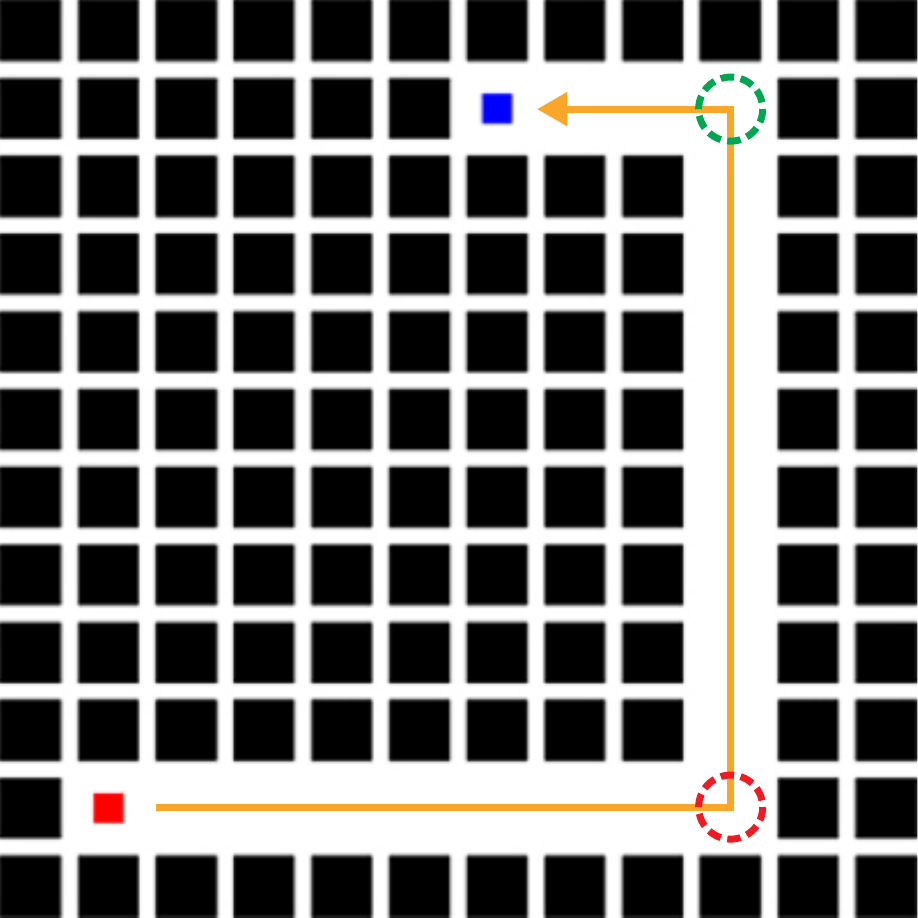}
        \caption{}
    \end{subfigure}%
    \\%
    \begin{subfigure}[b]{0.26\linewidth}
        \centering
        \includegraphics[width=\linewidth]{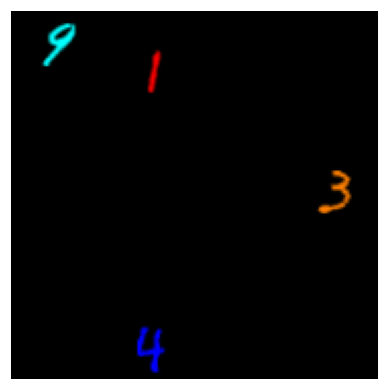}
        \caption{}
    \end{subfigure}%
    \begin{subfigure}[b]{0.26\linewidth}
        \centering
        \includegraphics[width=\linewidth]{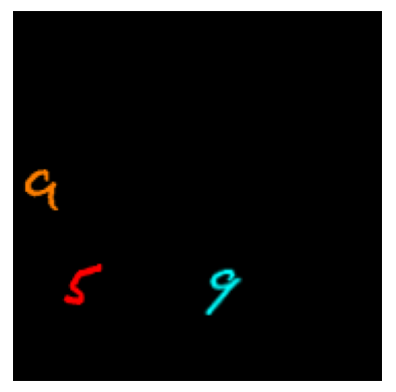}
        \caption{}
    \end{subfigure}%
    \begin{subfigure}[b]{0.26\linewidth}
        \centering
        \includegraphics[width=\linewidth]{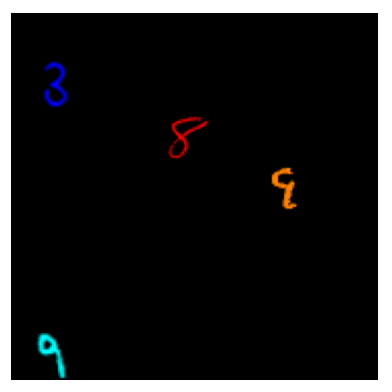}
        \caption{}
    \end{subfigure}%
    \caption{Examples of mazes and visual panels. (a) (b) Mazes have valid optimal paths in \ac{halma}, highlighted in green and blue. (c) A maze configuration leads to ambiguous observations, highlighted by a red circle at the bottom-right corner. (d) (e) Visual panels correspond to the corner highlighted in blue in (a) (b), respectively. (f) Visual panels corresponds to the corner, highlighted in red in (c).}
    \label{fig:appx:opt_path_validity}
\end{figure}

The validity of optimal paths in HALMA is defined to prevent the occurrence of \emph{ambiguous} states which may hinder the formation of strategies that are consistent within the problem domain of HALMA. In our design, the optimal strategies of all maze problems are expected to follow a \emph{meta-strategy}: making \emph{affordable} moves towards the direction of the goal. Unfortunately, such a requirement cannot be fulfilled by common methods for maze generation. 

Consider a common method to generate mazes in the grid-world: we first use randomized Prim's algorithm to create a connected area in the grid, and then decide the positions of initial state and goal state, which naturally produce an optimal path between them. \cref{fig:appx:opt_path_validity} shows three simple mazes that are generated by this method. In the mazes in \cref{fig:appx:opt_path_validity} (a) (b) agents who follow the meta-strategy above can indeed reach the goal. For example, at the bottom-right in \cref{fig:appx:opt_path_validity} (a), the agent may observe visual panels as in \cref{fig:appx:opt_path_validity} (d), wherein a $\mnist{1}{0}$ and a $\mnist{3}{1}$ indicate that there are walls 1-grid away \textbf{leftwards} $\caretl$ and 3-grid away \textbf{upwards} $\caretu$. The agent can also know from $\mnist{4}{5}$ and $\mnist{9}{4}$ that the goal state is 4-grid away on the \textbf{right} $\caretr$ and 9-grid away \textbf{upwards} $\caretu$. Obviously, the direction that is both with affordable moves and towards the goal is \textbf{upwards} $\caretu$. And it is also obvious that moving $\caretu$ can indeed reach the goal. This is also the case for all corners highlighted in green or blue circles in \cref{fig:appx:opt_path_validity} (a) (b). However, there is a state with ambiguous observation, highlighted by red circle at the bottom-right, in the maze in \cref{fig:appx:opt_path_validity} (c), wherein the agent may observe a visual panel depicted in \cref{fig:appx:opt_path_validity} (f). This visual panel contains a $\mnist{8}{0}$ and a $\mnist{9}{1}$, indicating that there are walls 8-grid away to the \textbf{left} $\caretl$ and 9-grid away \textbf{upwards} $\caretu$. The visual panel also includes a $\mnist{3}{5}$ and a $\mnist{9}{4}$, indicating that the goal state is 3-grid away to the \textbf{left} $\caretl$ and 9-grid away \textbf{upwards} $\caretu$. Both $\caretl$ and $\caretu$ are good candidate directions that are both with affordable moves and towards the goal. However, the global map tells us only $\caretu$ can lead the agent to the goal.

To eliminate the aforementioned ambiguity, in contrast to first generating the complete maze and then producing the optimal path, our solution is to first generate the \emph{valid} optimal path that rules out the ambiguity and then add deceptive branches to construct a grid-world maze. Formally, a path is considered \emph{invalid} if an agent possessing an oracle understanding of the concept space and acting in accord to the above {meta-strategy} fails to make decisions that lead the agent to the goal. We find that valid optimal paths are typically `L'-shaped from the initial state to the goal (see \cref{fig:appx:opt_path_validity} (a) (b)), whereas invalid paths are commonly `C'-shaped. In the latter, there is always a corner where the observation is ambiguous. In short, moving 1 unit on the \textit{valid optimal path} from the initial state position to the goal state position should reduce the Manhattan distance to the goal state position by 1.

\subsection{Generating Maze Problems}

\begin{figure}[th!]
    \centering
    \begin{subfigure}[b]{0.5\linewidth}
        \centering
        \includegraphics[width=0.8\linewidth]{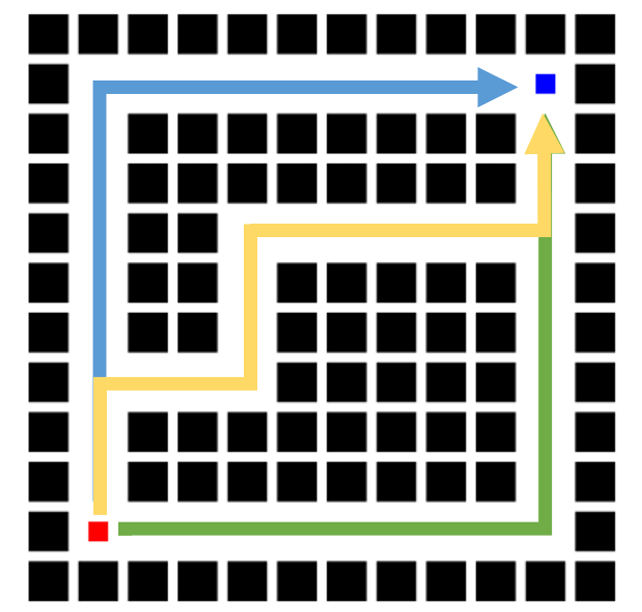}
        \caption{}
    \end{subfigure}%
    \begin{subfigure}[b]{0.5\linewidth}
        \centering
        \includegraphics[width=0.8\linewidth]{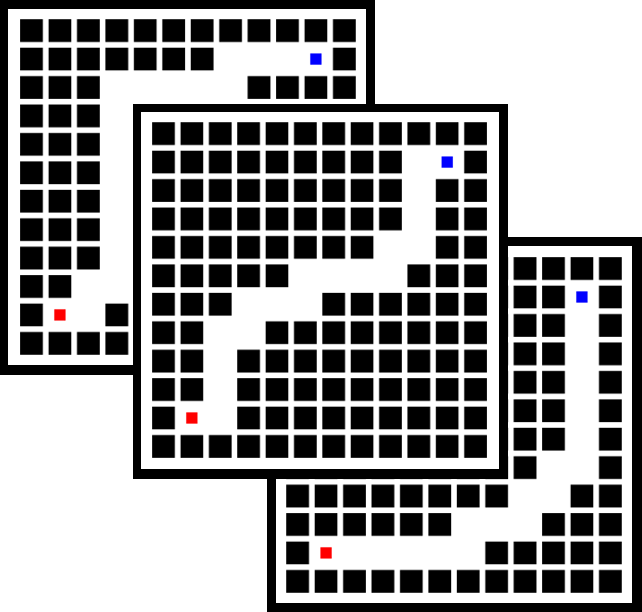}
        \caption{}
    \end{subfigure}%
    \caption{Illustration of valid optimal path generation. Given the initial state position and the goal state position, one can determine the directions where the optimal path should expand towards. For example, (a) Based on the initial state position, the optimal path can only expand upwards or to the right to reach the goal state position. (b) Examples of possible valid optimal paths.}
    \label{fig:appx:opt_path_gen}
\end{figure}

After clarifying the validity of optimal paths, we are able to build a pipeline to automatically generate the desired mazes. Assuming that the position of the initial state is on the bottom-left to the position of the goal state (see an example in \cref{fig:appx:opt_path_gen}), the optimal path should only expand upwards or to the right to reach the goal state position. Hence, given the horizontal offset $m$ and vertical offset $n$ from the initial state position to the goal state position, there should be $C(m+n-2,n-1)$ valid optimal paths in total. Note that in \ac{halma}, although all the positions of the initial state and the goal state are restricted within a $10\times10$ grid, it is able to produce $738,980$ possible optimal paths, exhibiting a rich and immense problem space in \ac{halma}.

\begin{wrapfigure}{Rt!}{0.5\linewidth}
    \vspace{-12pt}
    \centering
    \begin{subfigure}[b]{0.48\linewidth}
        \centering
        \includegraphics[width=\linewidth]{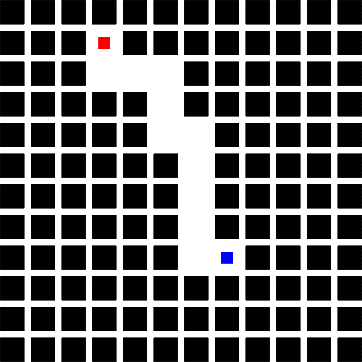}
        \caption{}
    \end{subfigure}%
    \hfill%
    \begin{subfigure}[b]{0.48\linewidth}
        \centering
        \includegraphics[width=\linewidth]{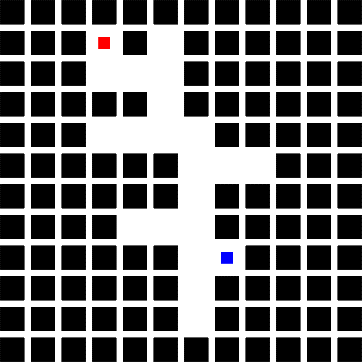}
        \caption{}
    \end{subfigure}%
    \caption{An example of adding deceptive branches to the valid optimal path.}
    \label{fig:appx:opt_path_w_branch}
\end{wrapfigure}

Next, we uniformly sample the optimal path from the maze set and add deceptive branches to these optimal paths. To maintain the validity of optimal path, we add a hint (\ie, ${\vcenter{\hbox{\includegraphics[width=0.03\linewidth]{hint/square}}}}, {\vcenter{\hbox{\includegraphics[width=0.03\linewidth]{hint/circle}}}}, {\vcenter{\hbox{\includegraphics[width=0.03\linewidth]{hint/triangle}}}}$, or ${\vcenter{\hbox{\includegraphics[width=0.03\linewidth]{hint/pentagon}}}}$) at each T-junction and crossing to indicate the direction the agent should move towards. In theory, the deceptive branches can be arbitrarily complex as they do not influence the validity of the optimal path. To test whether an agent understands the concept of these hints and successfully transfers the learned knowledge to novel problems, we set the average \emph{depth} of deceptive branches to $2$ in the training set and $5$ in the testing set. To provide sufficient training data for an agent to recognize these hints, we set the average \emph{branching number} to $5$ in the training set.

\subsection{An Example Trial}

\begin{figure}[th!]
    \centering
    \begin{subfigure}[b]{0.25\linewidth}
        \centering
        \includegraphics[width=\linewidth]{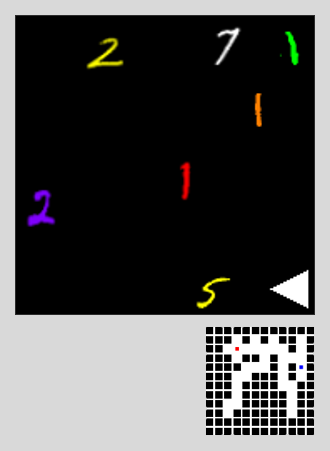}
        \caption{step 1}
    \end{subfigure}%
    \begin{subfigure}[b]{0.25\linewidth}
        \centering
        \includegraphics[width=\linewidth]{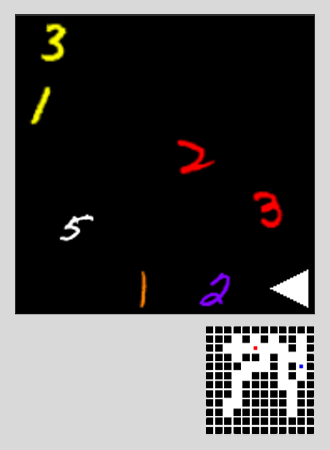}
        \caption{step 2}
    \end{subfigure}%
    \begin{subfigure}[b]{0.25\linewidth}
        \centering
        \includegraphics[width=\linewidth]{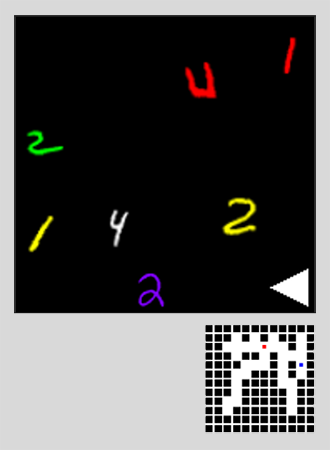}
        \caption{step 3}
    \end{subfigure}%
    \begin{subfigure}[b]{0.25\linewidth}
        \centering
        \includegraphics[width=\linewidth]{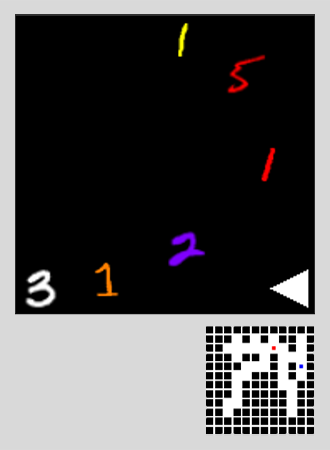}
        \caption{step 4}
    \end{subfigure}%
    \\
    \begin{subfigure}[b]{0.25\linewidth}
        \centering
        \includegraphics[width=\linewidth]{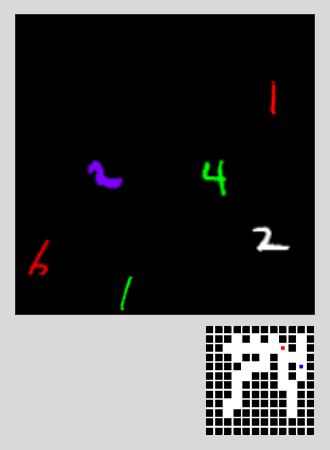}
        \caption{step 5}
    \end{subfigure}%
    \begin{subfigure}[b]{0.25\linewidth}
        \centering
        \includegraphics[width=\linewidth]{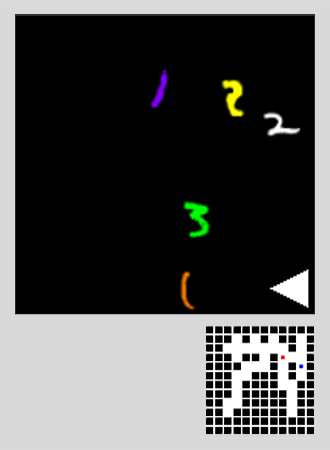}
        \caption{step 6}
    \end{subfigure}%
    \begin{subfigure}[b]{0.25\linewidth}
        \centering
        \includegraphics[width=\linewidth]{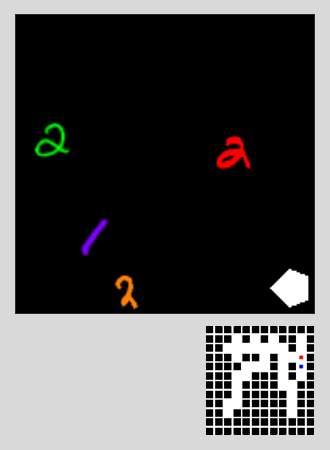}
        \caption{step 7}
    \end{subfigure}%
    \begin{subfigure}[b]{0.25\linewidth}
        \centering
        \includegraphics[width=\linewidth]{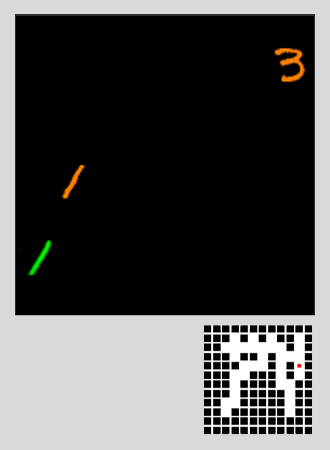}
        \caption{step 8}
    \end{subfigure}%
    \caption{Visualization of an example trial completed by the oracle agent in $8$ steps. Mazes at the bottom-right in (a)-(h) illustrate the trajectory of the oracle agent.}
    \label{fig:appx:example_trial}
\end{figure}

In this section, we visualize an example trial completed by the \emph{oracle} agent to further illustrate \ac{halma}. The maze of this example is the same as the one we present in our interactive website \url{http://halma-proj.github.io/}. So we strongly recommend you to visit this website for a grounded experience when you read this subsection. Since this example trial is from an oracle agent, it is optimal in terms of exploration efficiency: it is finished in $8$ moves; consecutive frames are shown in \cref{fig:appx:example_trial}. Below, we provide detailed explanation of how the oracle agent makes its decision at each timestep:
\begin{itemize}[noitemsep,nolistsep]
    \item[(a)] The oracle agent is spawned at an initial state position, highlighted by the red dot in the maze panel in \cref{fig:appx:example_trial} (a). Its observation is the visual panel, consisting of MNIST digits and a $\Triangle$ hint. Recall that the ground-truth semantics of $\Triangle$ indicates that the agent should move to the right, \ie, $\caretr$. Therefore, the agent who understands the meaning of $\Triangle$ would only need to know the distance to the wall and to the nearest T-junction or crossing\footnote{We will use the term \emph{crossing} to refer to either of them henceforth, as well as in the main text.} to the right in order to decide which move to take. Finally, recall that the $\texttt{yellow}$ color is connected with $\caretr$; the agent needs to make a comparison between the $\mnist{2}{2}$ and the $\mnist{5}{2}$, and chooses the lesser digit (\ie, 2) as the distance it moves $\caretr$. So the optimal move at this frame is $\caretr:\dicetwo$ or $\caretr:\diceone+\diceone$.
    \item[(b)-(d)] In these frames, the oracle agent takes similar moves as in \cref{fig:appx:example_trial} (a). That is, the oracle agent chooses the move $\caretr:\diceone$ in these three frames. Note that in \cref{fig:appx:example_trial} (d), there is only one $\texttt{yellow}$ color MNIST digit (\ie, $\mnist{1}{2}$) in the visual panel; therefore the agent may not need to make comparisons between digits. Hints $\Triangle$ appear in all these visual panels since the oracle agent always stops at crossings.
    \item[(e)] The oracle agent does not observe any hints for direction (\ie, $\{\Square, \Circle, \Triangle, \Pentagon\}$) in the visual panel because it is not at a crossing, therefore it needs to reason from the observation for the direction of the goal. The $\mnist{2}{6}$ and the $\texttt{white}$ digit 2 indicate that the goal position is 2-grid downwards and 2-grid to the right. Additionally, the agent also observes no $\texttt{yellow}$ digit in the panel, which indicates that the the grid direct to the agent's right is a wall hence $\caretr$ is not a valid direction. Therefore, the agent should move downwards (\ie $\caretd$). Finally, recall that the $\texttt{green}$ color is connected with $\caretd$; the agent needs to make a comparison between the $\mnist{1}{3}$ and the $\mnist{4}{3}$, and chooses the lesser digit (\ie, 1) as the distance it moves downwards. So the optimal move at this frame is $\caretd:\diceone$.
    \item[(f)-(g)] The rationale of optimal moves from the oracle agent is the same as in previous frames: In frame (f) the hint is $\Triangle$, so the agent should move rightwards (\ie $\caretr$). There is only one digit $\mnist{2}{2}$ connected to this direction, so the optimal move is $\caretr:\dicetwo$. In frame (g), the hint is $\Pentagon$, which indicates that the agent should move downwards (\ie, $\caretd$). Additionally, the $\mnist{1}{6}$ and $\mnist{2}{3}$ indicate that the goal state position is 1-grid downwards, and there is no obstacle in the way until 2-grid away. The oracle agent can infer that it should move downwards $\caretd$ by 1 unit to reach the goal. So the optimal move at this frame is $\caretd:\diceone$. 
    \item[(h)] This frame shows the goal state in this trial. The example trial ends at this frame.
\end{itemize}
After reaching the goal in the aforementioned trial, the agent will be respawned to the starting position due to our formulation of Rapid Problem-Solving. And the agent is expected to merge the four consecutive right moves in frame (a)-(d) into $\caretr:\dicethree+\dicetwo$ or $\caretr:\dicethree+\diceone+\diceone$ or $\caretr:\diceone+\diceone+\diceone+\dicetwo$ or their equivalents to reduce the total timesteps by 3. 

As you can see, the rationale of picking the optimal moves, \ie the \emph{meta-strategy}, is consistent in all frames: (i) if the agent is at the crossing, the agent should move towards the direction indicated by the hint, (ii) if the agent is not at the crossing, the agent should move towards the goal direction, (iii) the total number of units to move at each frame depends on the MNIST digits whose color aligns with the valid direction, (iv) if it is the first trial in a maze, the agent should always stop at the crossing to obtain the hint. Such a consistency is only possible with our definition of \textit{valid optimal paths}. 

\section{Formal Definitions of Concept Spaces}
\label{appx:concept_space}

\subsection{Preliminary}

For the sake of formalism, we borrow the terminology from the \emph{General Pattern Theory} \citep{grenander1993general}. In case readers are not familiar with the General Pattern Theory, it is a mathematical study of \emph{regular structures} --- \emph{configuration spaces, patterns} to account for the \emph{combinatory} principle of our world. Adopting the language of abstract algebra, Grenander calls the basic unit of a regular structure/configuration space a \emph{generator}, generically denoted as $g_i$. Any $g_i$ is associated with a number of \emph{bonds} $\beta_j$, whose \emph{value} $\beta_j(g_i)$ shall be within the \emph{bond value space} $B$. Generators are combined together by \emph{connectors}. A connector $\sigma$ is a graph, say with $n$ sites. When $n$ generators are placed on a connectors' sites, we have a \emph{configuration}, $c=\sigma (g_1, g_2, ..., g_n)$, which comes together with a set of \emph{bond relations} $\rho:B\times B\rightarrow\{\texttt{TRUE}, \texttt{FALSE}\}$. A configuration is called \emph{regular} if all bond relations return $\texttt{TRUE}$. 

Despite of its generality, the formal language used by Grenander might appear somewhat abstract or peculiar to researchers in our community. Hence, we further elaborate below, from the perspective of \emph{grammar}. A grammar is a regular structure, mostly studied in the community of natural language or linguistics to elucidate the combinatorial expressiveness in generating an immense set of configurations by composing only a considerably smaller set of words, using production rules. To account for the similar compositional and hierarchical nature in visual scenes, \citet{zhu2007stochastic} introduced a stochastic grammar to the community of vision. They proposed an image grammar in an \emph{And–Or Graph} (AOG) representation, where each Or-node points to alternative sub-configurations, and each And-node is decomposed into a number of sub-components. An AOG represents (i) the hierarchical decompositions from scenes to primitives and pixels, via non-terminal and terminal nodes, and (ii) the contexts for spatial and functional relations by horizontal links among the nodes. Below, to make this appendix self-contained, we summarize some key definitions:

\theoremstyle{definition}
\begin{definition}[Vocabulary]
The vocabulary $V$ is a set of generators $g_i(\alpha_i)$, each associated with its
bonds, $\beta_i = (\beta_{i,1},...,\beta_{i,d(i)})$. $\alpha_i$ is a vector of \emph{attributes}. For instance, a visual generator may contain material properties of an object or the gender of a person as its attributes.\footnote{In computer vision, attributes are some properties of objects or agents that tend to remain the same.} Bonds need to be connected with other bonds to form \emph{attributed relations}; see the next definition.
\end{definition}

\theoremstyle{definition}
\begin{definition}[Attributed Relations]
Given an arbitrary set of generators $V$, a binary relation is a subset of the product set $V\times V$
$$\{(u, v)\}\subset V \times V.$$
An attributed binary relation is an augmented binary relation with a vector of attributes $\sigma$ and $\rho$
$$E = \{(u, v;\sigma, \rho) : u, v \in V\},$$
where $\sigma(u, v)$ represents the \emph{connector} that binds $u$ and $v$, and $\rho(s, t)$ is a real number measuring the compatibility between $u$ and $v$. Then $\langle V, E\rangle$ is a graph, expressing the generalized relation $E$ on $S$. It is the \emph{relation} that you are familiar with in object-oriented language such as First-Order Logics. For instance, the distance between two objects is an attributed relation. A $k$-way attributed relation is defined in a similar way as a subset of $V^k$.
\end{definition}

\theoremstyle{definition}
\begin{definition}[Configuration]
A configuration $C$ is a one-layer graph, often flattened from its hierarchical representation
$$C = \langle V, E \rangle.$$
For a visual scene, it is a spatial layout of entities in a scene at certain level of abstraction.
\end{definition}

\theoremstyle{definition}
\begin{definition}[Parse Graph]
A \emph{parse graph} $pg$ consists of a hierarchical \emph{parse tree} (defining ``vertical'' edges) and a number of relations E (defining ``horizontal edges''):
$$pg = \langle pt, E \rangle.$$
The parse tree $pt$ is also an \emph{And-tree}, whose non-terminal nodes are all And-nodes. The decomposition of each And-node $A$ into its parts is given by a \emph{production rule}, which now produces not a string (like in natural language or linguistics) but a configuration:
$$\sigma : A \rightarrow C = \langle V, E \rangle.$$
A production should also associate the open bonds of $A$ with open bonds in $C$. The whole parse tree is a sequence of production rules:
$$pt = (\sigma_1, \sigma_2, ... \sigma_n).$$
The horizontal links $E$ consists of a number of directed or undirected relations among the \emph{terminal} or \emph{non-terminal} nodes:
$$E = E_{r_1} \cup E_{r_2} \cup ... \cup E_{r_k}.$$
These relations can be spatial relations, semantic relations, affordance relations, and causal relations. 
A parse graph $pg$, when \emph{collapsed}, produces a series of flat configurations at each level of abstraction/detail:
$$pg \implies C.$$
\end{definition}

\setstretch{0.99}

\theoremstyle{definition}
\begin{definition}[And-Or Graph]
An And–Or Graph is a 6-tuple for representing an grammar $\mathcal{G}$.
$$G = \langle S, V_N, V_T, \mathcal{R}, \Sigma, \mathcal{P} \rangle.$$
$S$ is the root node of a scene, $V_N = V^{\text{and}} \cup V^{\text{or}}$ is a set of non-terminal nodes, including an And-node set $V^{and}$ and an Or-node set $V^{or}$. The And-nodes plus sub-graphs formed by their children are the productions, whereas the Or-nodes are the vocabulary items.
$V_T$ is a set of terminal nodes, for instance, visual primitives, parts, and objects. $\mathcal{R}$ is a number of relations between the nodes, $\Sigma$ is the set of all valid/regular configurations derivable from the grammar, \ie, its language. $\mathcal{P}$ is the probability model defined on the And–Or Graph.
\end{definition}

In sum, as a generic representation, an And-Or Graph can represent the hierarchical and relational knowledge of a visual scenario.\footnote{Note that by \emph{representation}, we do not necessarily mean how an artificial agent should represent such knowledge. Rather, it is a formalism for us humans to understand the internal structure of \ac{halma}.} In the following subsections, we concretely define the \emph{configuration space} of \ac{halma} by grounding abstract notions in this subsection to specific components. 

\subsection{Concept Spaces of \texorpdfstring{\ac{halma}}{}}

\theoremstyle{definition}
\begin{definition}[Axioms for Equivalence Relation $=$]
An equivalence relation is a binary relation that is reflexive, symmetric, and transitive.
For any generators $g_1$, $g_2$, and $g_3$:
\begin{itemize}[noitemsep,nolistsep]
    \item $g_1 = g_1$, (\emph{Reflexivity})
    \item $g_1 = g_2$ if and only if $g_2=g_1$, (\emph{Symmetry})
    \item if $g_1 = g_2$ and $g_2 = g_3$, then $g_1 = g_3$. (\emph{Transitivity})
\end{itemize}
\end{definition}

\theoremstyle{definition}
\begin{definition}[Axioms for Partial Order Relation $\leq$]
A partial order is a binary relation that is reflexive, antisymmetric, and transitive.
For any generators $g_1$, $g_2$, and $g_3$:
\begin{itemize}[noitemsep,nolistsep]
    \item $g_1 \leq g_1$, (\emph{Reflexivity})
    \item if $g_1 \leq g_2$ and $g_2 \leq g_1$, then $g_1=g_2$, (\emph{Antisymmetry})
    \item if $g_1 \leq g_2$ and $g_2 \leq g_3$, then $g_1 \leq g_3$. (\emph{Transitivity})
\end{itemize}
\end{definition}

\theoremstyle{definition}
\begin{definition}[Addition on Nature Numbers $+$]
Given $\mathbb{N}$ and its \emph{successor} function $s$ by \emph{Peano Axioms}, we may have a \emph{group} $\langle\mathbb{N},+\rangle$ if we define \emph{addition} $+$ as: for $n, m\in\mathbb{N}$,
\begin{itemize}[noitemsep,nolistsep]
    \item $n+0=n$,
    \item $n+s(m) = s(n+m)$.
\end{itemize}
\end{definition}

\theoremstyle{definition}
\begin{definition}[Subtraction on Nature Numbers $-$] Given $\mathbb{N}$ and $\leq$,
\begin{itemize}[noitemsep,nolistsep]
    \item let $m, n\in\mathbb{N}$, such that $m\leq n$;
    \item let $p\in\mathbb{N}$, such that $n=m+p$. 
\end{itemize}
We define \emph{subtraction} $-$ as $n-m=p.$
\end{definition}

\theoremstyle{definition}
\begin{definition}[Spatial Grammar of \ac{halma}]
The spatial grammar of \ac{halma} is an Spatial And–Or Graph (S-AOG), which is a 6-tuple 
$$G_S = \langle S_S,V_{N_S}, V_{T_S}, \mathcal{R}_S, \Sigma_S, \mathcal{P}_S\rangle,$$
where $S_S$ is the root node that represents the set of all visual panels, thus an Or-Node connected to nodes in $V_{N_S}$. There is only one element $v$ in $V_{N_S}$, representing an instance of visual panel. $v$ is a Set-Node since the number of digits in the panel may vary with different state; recall that it is because \emph{zero} does not appear in the panel. $v$ produces all MNIST digits $\texttt{d}_\texttt{i}$ (or hints) in the panel; it is a \emph{composed concept}. These MNIST digits consist the terminal node $V_{T_S}$. They are \emph{attributed} with \texttt{color}, \texttt{scale}, \texttt{location}, \texttt{indication}, and \texttt{category}. Specifically, $\texttt{color}=\{\texttt{red}, \texttt{orange}, \texttt{yellow}, \texttt{green}, \texttt{cyan}, \texttt{blue}, \texttt{purple}, \texttt{white}\}$, and $\texttt{indication}=\{\texttt{wall}\lor \texttt{crossing}, \texttt{goal}\}$. Ideally, the visual panel contains all nature numbers, $\texttt{category}^*=\mathbb{N}\cup\{\Circle,\Triangle,\Square,\Pentagon\}$. Currently, however, we only consider $\texttt{category}=\{\texttt{1}, \texttt{2}, \texttt{3}, \texttt{4}, \texttt{5}, \texttt{6}, \texttt{7}, \texttt{8}, \texttt{9}\}\cup\{\Circle,\Triangle,\Square,\Pentagon\}$. There is a \emph{bijection} between $\texttt{color}$ and $\{\caretu,\caretd,\caretl,\caretr\}\times\texttt{indication}$, which gives rise to a \emph{partition}, $\texttt{type}$, over $V_{T_S}$. As terminal nodes, $V_{T_S}$ are \emph{atomic generators}, hence \emph{primitive concepts}. Though there can be many possible relations between these generators (\eg, distance between MNIST digits, ordering of \texttt{scale} between MNIST digits), only the (\emph{strict}) \emph{partial order} over $\texttt{category}\times\{\caretu,\caretd,\caretl,\caretr\}$, \ie, $\langle\texttt{S}, <\rangle$ is crucial to the task of \ac{halma}. The definition of $\langle\texttt{S}, <\rangle$ would come clear once we define $\texttt{S}$ and how the concept of $\mathbb{N}$ is \emph{bootstrapped} and grounded to $V_{T_S}$. $\mathcal{P}$ depends on the underlying maze problem since the valid configuration space $\Sigma_S$ of this grammar is all descriptions of states. 
\end{definition}

\setstretch{0.98}

\theoremstyle{definition}
\begin{definition}[Semantics in \ac{halma}]
The \emph{semantics} $\texttt{S}$ in \ac{halma} is a relation, a subset of $V_{T_S} \times \texttt{category}\times\{\caretu,\caretd,\caretl,\caretr\}$
$$\texttt{S}\subset V_{T_S} \times \texttt{category}\times\{\caretu,\caretd,\caretl,\caretr\},$$
which is the ground-truth labeling of MNIST digits and their colors. For simplicity, we would slighly abuse this notion: In the remainder of the paper, we may regard $\texttt{S}$ as a function $V_{T_S} \rightarrow \texttt{category}\times\{\caretu,\caretd,\caretl,\caretr\}$ and also regard it as the \emph{range} of this function.
\end{definition}

\theoremstyle{definition}
\begin{definition}[Temporal Grammar of \ac{halma}]
The temporal grammar of \ac{halma} is a Temporal And–Or Graph (T-AOG), which is a 6-tuple 
$$G_T = \langle S_T, V_{N_T}, V_{T_T}, \mathcal{R}_T, \Sigma_T, \mathcal{P}_T \rangle,$$
where $S_T$ is the root node that represents the set of all options, thus an Or-Node connected to elements in $V_{N_T}$. Different from the spatial grammar, the temporal grammar has richer hierarchical structure, therefore there are more than one element in $V_{N_T}$, each representing an option $\texttt{opt}$. An option is a \emph{composed concept}, which produces its constituting options/actions. The production rule $\rho$ is defined by the operation $+$. Production terminates when reaching terminal nodes $V_{T_T}=\{\dicez,\diceone,\dicetwo,\dicethree\}$. Since each of them are mapped to a semantic meaning (\ie, moving 0, 1, 2, or 3 units), they are \emph{primitive concepts} of this grammar. All actions and options are attributed with $\{\caretu,\caretd,\caretl,\caretr\}$, which regularizes the production to be within the same \texttt{type}. Ideally, if we could build maze with infinite size, for each \texttt{type}, the production rule would specify a \emph{group} over all nature numbers $\langle\mathbb{N}, +\rangle$. With that said, the only element in $\mathcal{R}_T$ is \emph{equality} $=$. If we represent all elements in $\langle\mathbb{N}, +\rangle$ with sequences of primitive set along with equality over them, we have the valid configuration space $\Sigma_T$. $\mathcal{P}$ is the prior distribution of this numerical decomposition. 
\end{definition}

\theoremstyle{definition}
\begin{definition}[Affordance in \ac{halma}]
The \emph{affordance} $\texttt{A}$ in \ac{halma} is a relation, a subset of $V_{T_S} \times (V_{N_T} \cup V_{T_T})$
$$\texttt{A} \subset V_{T_S} \times (V_{N_T} \cup V_{T_T}),$$
which is a partial $\geq$ relation between the semantics of atomic generators in the spatial grammar and all generators in the temporal grammar. It is a partial relation because defined within each \texttt{type} and its \emph{inverse}. Namely, it is defined based on $\langle\mathbb{N}, +/-, \leq\rangle$. An action/option is \emph{affordable} in a state if this relation returns true. Hence, affordance is a \emph{bootstrapped concept} emerged from agents' interaction with the environment. Recall that there may be two MNIST digits with the same color in one panel; the lesser one indicates the distance till the nearest crossing, and the greater one indicates the distance to the wall. Regardless of their difference in semantics, both of them fit this definition well, though only the greater digit indicates the ground-truth affordance in the current maze.
\end{definition}

\theoremstyle{definition}
\begin{definition}[Causal Structure of \ac{halma}]
The causal structure of \ac{halma} is a Causal And–Or Graph (C-AOG), which is a 6-tuple 
$$G_C = \langle S_C, V_{N_C}, V_{T_C}, \mathcal{R}_C, \Sigma_C, \mathcal{P}_C \rangle,$$
where $S_C$ is the root node that represents the set of all scenarios, thus an Or-Node connected to elements in $V_{N_C}$. $G_C$ links $G_S$ and $G_T$ together. Since the environment of \ac{halma} is Markovian, we have $\Sigma_C\subset\Sigma_S\times\Sigma_T\times\Sigma_S$.\footnote{Otherwise, causal configurations would be non-Markovian, $\Sigma_C\subset(\Sigma_S\times\Sigma_T)^*\times\Sigma_S$.} With that said, \emph{generators} in the causal structure include $V_{N_S}\cup V_{T_S}$ and $V_{N_T}\cup V_{T_T}$. Namely, $V_{N_C} = V_{N_S}\cup V_{N_T}$; $V_{T_C} = V_{T_S}\cup V_{T_T}$. Definitions of production rules in the causal structure inherit from the spatial grammar and the temporal grammar. What uniquely defined here is $\mathcal{R}_C=\{\langle\texttt{S}, <\rangle\, \langle\texttt{A}, +/-. \leq\rangle, \langle\texttt{C}, +/-, =\rangle\}$. All these three relations are derivable from $\langle\mathbb{N}, +/-, =, <\rangle$. In the current setup of \ac{halma}, $\mathcal{P}_C$ is deterministic. Reader who are familiar with symbolic planning may find the similarity between $G_C$ and STRIPS-style action languages \citep{fikes1971strips}. Specifically, affordance $\texttt{A}$ corresponds to the \emph{precondition} of an action, whereas \emph{causality} corresponds to the \emph{effect} of an action, to be defined below.
\end{definition}

\theoremstyle{definition}
\begin{definition}[Causality in \ac{halma}]
The \emph{causality} $\texttt{C}$ in \ac{halma} is a relation, a subset of $V_{T_S} \times (V_{N_T}\cup V_{T_T})\times V_{T_S}$
$$\texttt{C}\subset V_{T_S} \times (V_{N_T}\cup V_{T_T})\times V_{T_S},$$
which is a partial $=$ relation between (i) the Cartesian product of the semantics of atomic generators in the spatial grammar and (ii) all generators in the temporal grammar and (iii) the semantics of atomic generators in the spatial grammar. Similar to the semantics in \ac{halma}, we would somewhat abuse its notion and refer to it as a function $V_{T_S} \times (V_{N_T}\cup V_{T_T})\rightarrow V_{T_S}$. Similar to $\texttt{A}$, it is a partial relation because defined within each \texttt{type} and its \emph{inverse}. For domains where it is defined, its definition is based on $\langle\mathbb{N}, +/-, =\rangle$. It is also a \emph{bootstrapped concept} emerged from interaction. 
\end{definition}

\setstretch{1}

\section{Statistics of Visual Panels}
\label{appx:stats}

\begin{figure}[th!]
    \centering
    \begin{subfigure}[b]{0.43\linewidth}
        \centering
        \includegraphics[width=\linewidth]{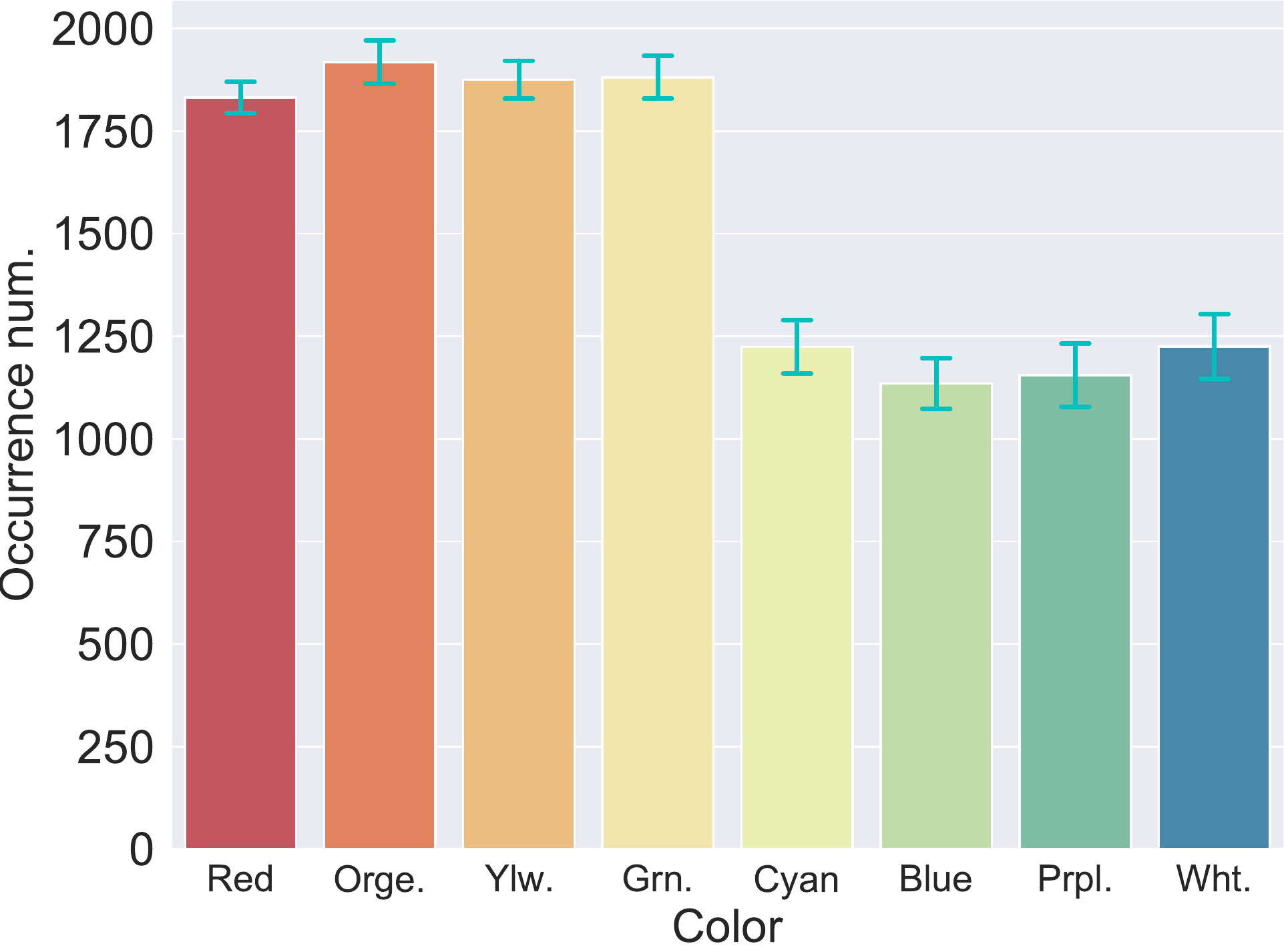}
        \caption{Colors}
    \end{subfigure}%
    \begin{subfigure}[b]{0.43\linewidth}
        \centering
        \includegraphics[width=\linewidth]{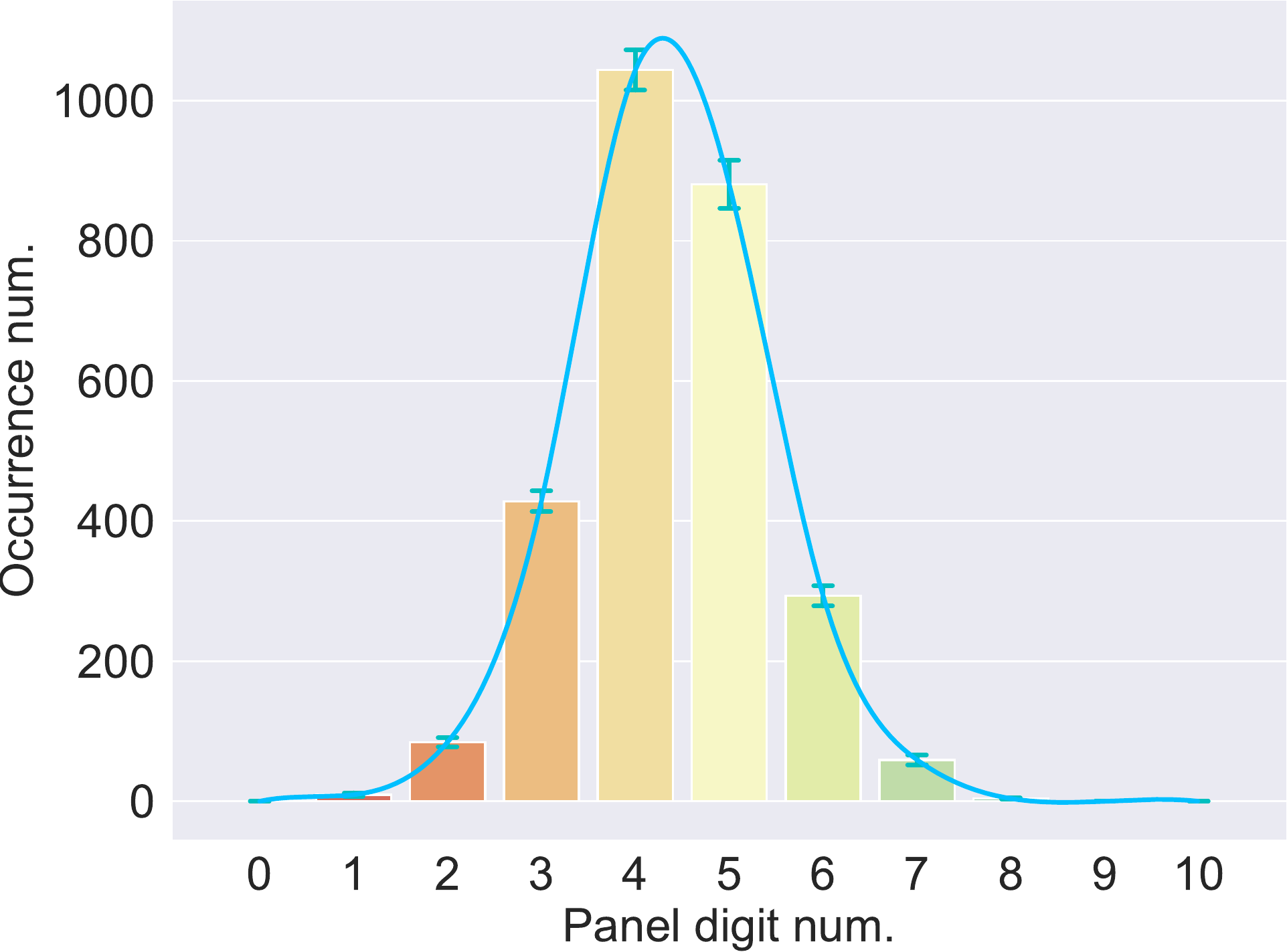}
        \caption{Number of digits in a panel}
    \end{subfigure}%
    \\
    \begin{subfigure}[b]{0.43\linewidth}
        \centering
        \includegraphics[width=\linewidth]{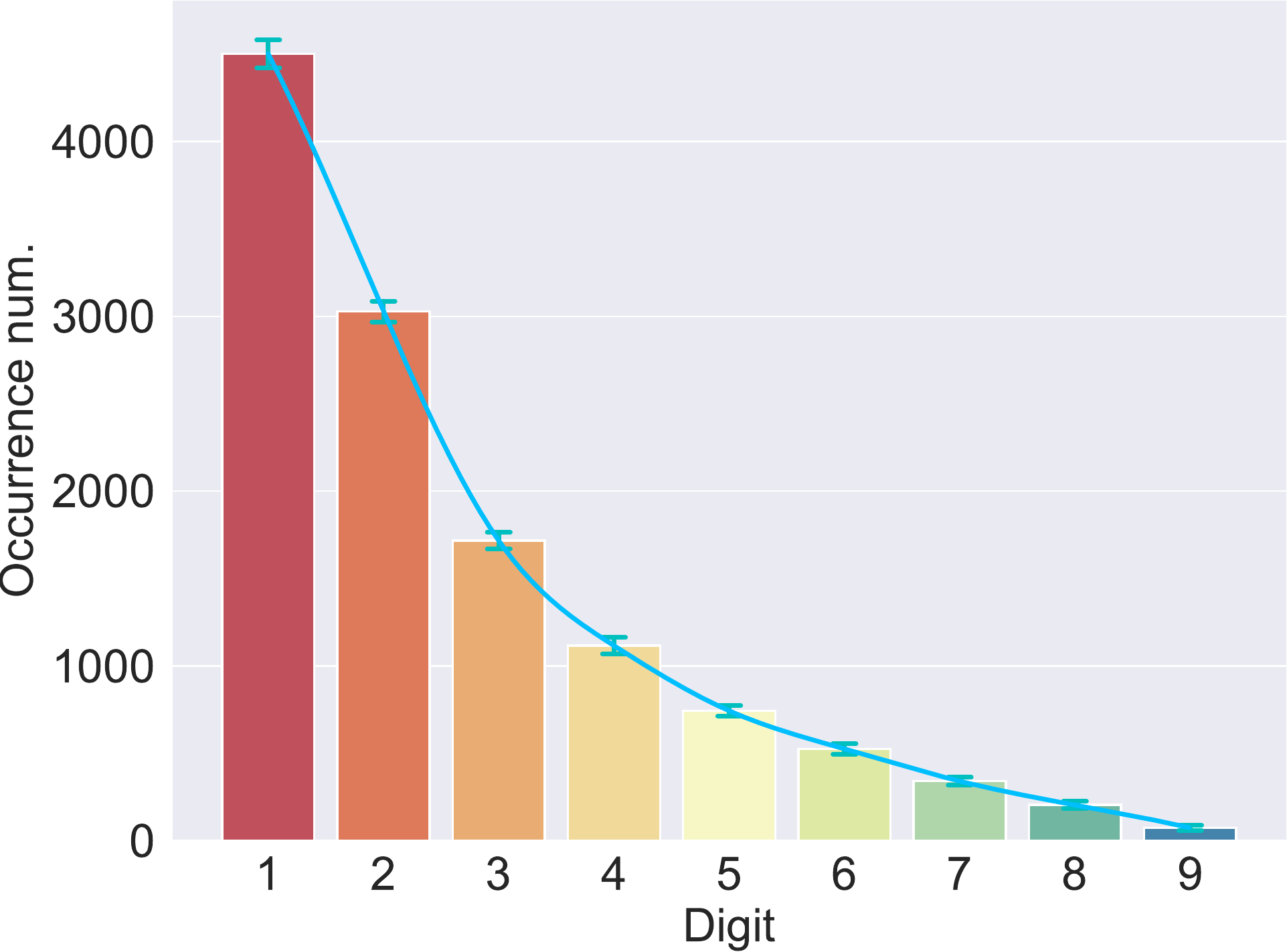}
        \caption{Digits distribution over 10 sets}
    \end{subfigure}%
    \begin{subfigure}[b]{0.43\linewidth}
        \centering
        \includegraphics[width=\linewidth]{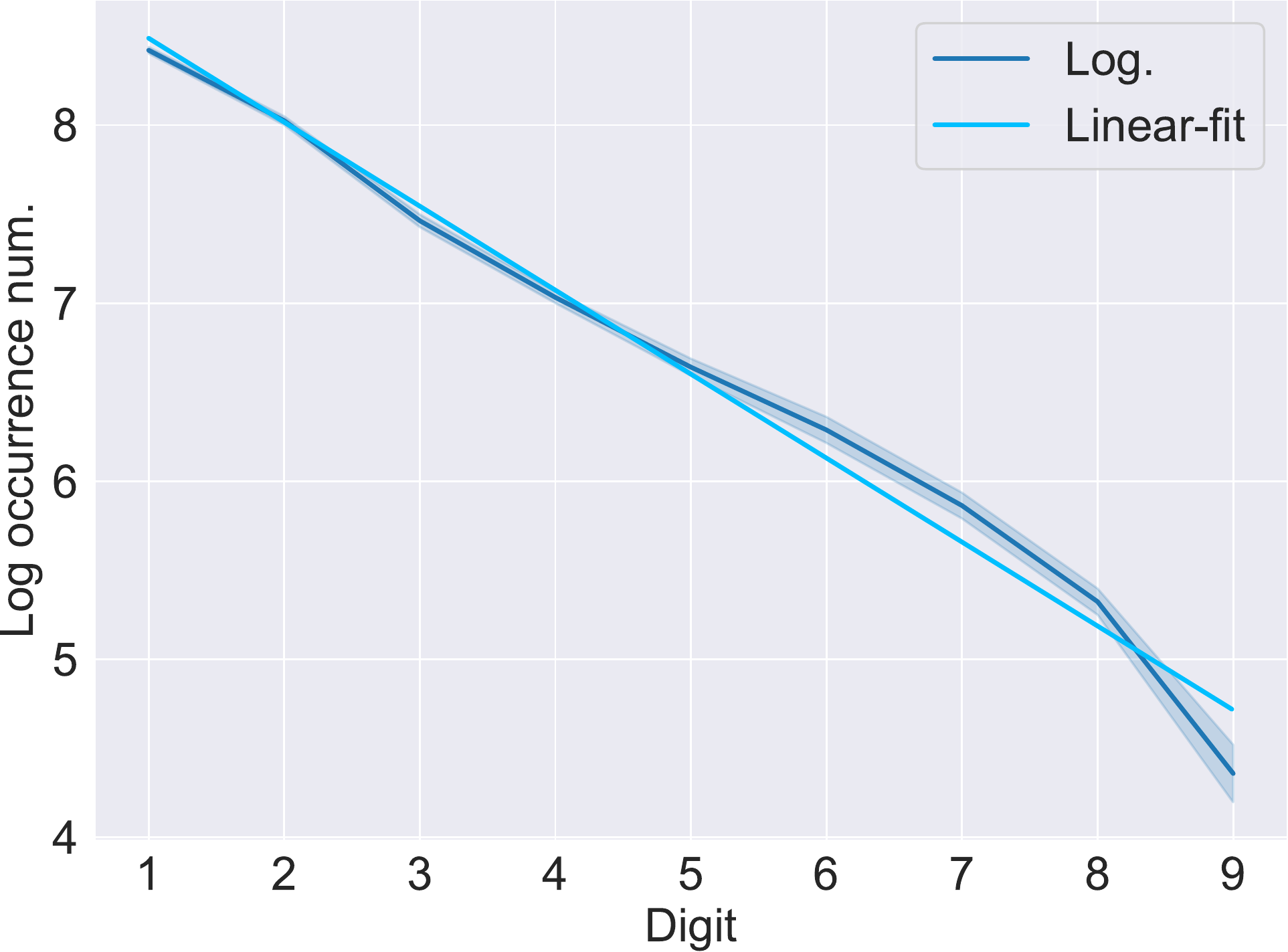}
        \caption{Digit distribution in log-scale}
    \end{subfigure}%
    \caption{Key statistics of visual panels in the \ac{halma} training set. Each training set contains 100 \ac{halma} grid-world mazes. We randomly sample 10 training sets and report the mean and standard deviation of the occurrence count of (a) colors, (b) number of digits in a panel, (c) digits distribution over these 10 sets, and (d) digits distribution in log-scale.}
    \label{fig:appx:maze_stats}
\end{figure}

Recall that in \ac{halma}, we use eight colors, \ie, \texttt{red}, \texttt{orange}, \texttt{yellow}, \texttt{green}, \texttt{cyan}, \texttt{blue}, \texttt{purple}, and \texttt{white}, to specify the type of digits. Digits indicating the distance till a wall or the nearest crossing towards each direction (\ie, $\caretl, \caretu, \caretr$, and $\caretd$) are colored \texttt{red}, \texttt{orange}, \texttt{yellow}, and \texttt{green}, respectively. Digits indicating the offset to the goal state are colored \texttt{cyan}, \texttt{blue}, \texttt{purple}, and \texttt{white}. Following the design of CLEVR \citep{johnson2017clevr}, in \ac{halma}, we deliberately control the distribution of visual attributes, especially of \texttt{COLOR}, by sightly adjusting generated mazes to form a uniform distribution of digit type. Such design help to avoid possible strong biases in the data that agents can exploit to correctly take actions without reasoning. Below, we report key statistics of visual panels in the \emph{training set} to demonstrate the uniformity of attributes distribution.

\cref{fig:appx:maze_stats} (a) illustrates the color distribution of the visual panels. We produce an approximately uniform distribution for the color connected with distance to walls and crossings (\ie, \texttt{red}, \texttt{orange}, \texttt{yellow}, and \texttt{green}) and for the color connected with offset to goal state (\ie, \texttt{cyan}, \texttt{blue}, \texttt{purple}, and \texttt{white}) separately. We uniformly sample optimal paths and add deceptive branches when creating the mazes in the training set (see details in  \cref{appx:halma_problems}) to form this distribution as an attempt to mitigate the color-conditional bias in the training set.

\cref{fig:appx:maze_stats} (b) shows the distribution of number of digits in the panel. Number of digits in the panel are in an unimodal distribution. More than 90\% panels in the training set has a number of digits between 3 and 6. Only $<$10\% panels have 1-2 or 7-8 digits. No panel has a number of digits greater than 8.

\cref{fig:appx:maze_stats} (c) (d) plot the distribution of digit in visual panels, revealing a long-tail distribution, where digit `1' has an occurrence number over 4,000, and digit `9' has an occurrence number less than 100. We consider this design as a nature of \ac{halma} training set. Note that a greater digit tends to co-occur with the lesser digits in \ac{halma}. For instance, if the agent passes a 9-grid-long passage step-by-step in a maze, it would observe not only the digit `9,' but also all the digits from `1' to `8.' Additionally, since we uniformly add branches on the optimal paths to create crossings, it adds to the occurrence of lesser digits. In essence, this almost log-linear distribution aligns well with the natural distribution of digits or words for numbers in human language \citep{dehaene1992cross}.

\setstretch{1}

\section{Dynamically Generate Generalization Tests}
\label{appx:kb}

\begin{figure}[th!]
    \includegraphics[width=\linewidth]{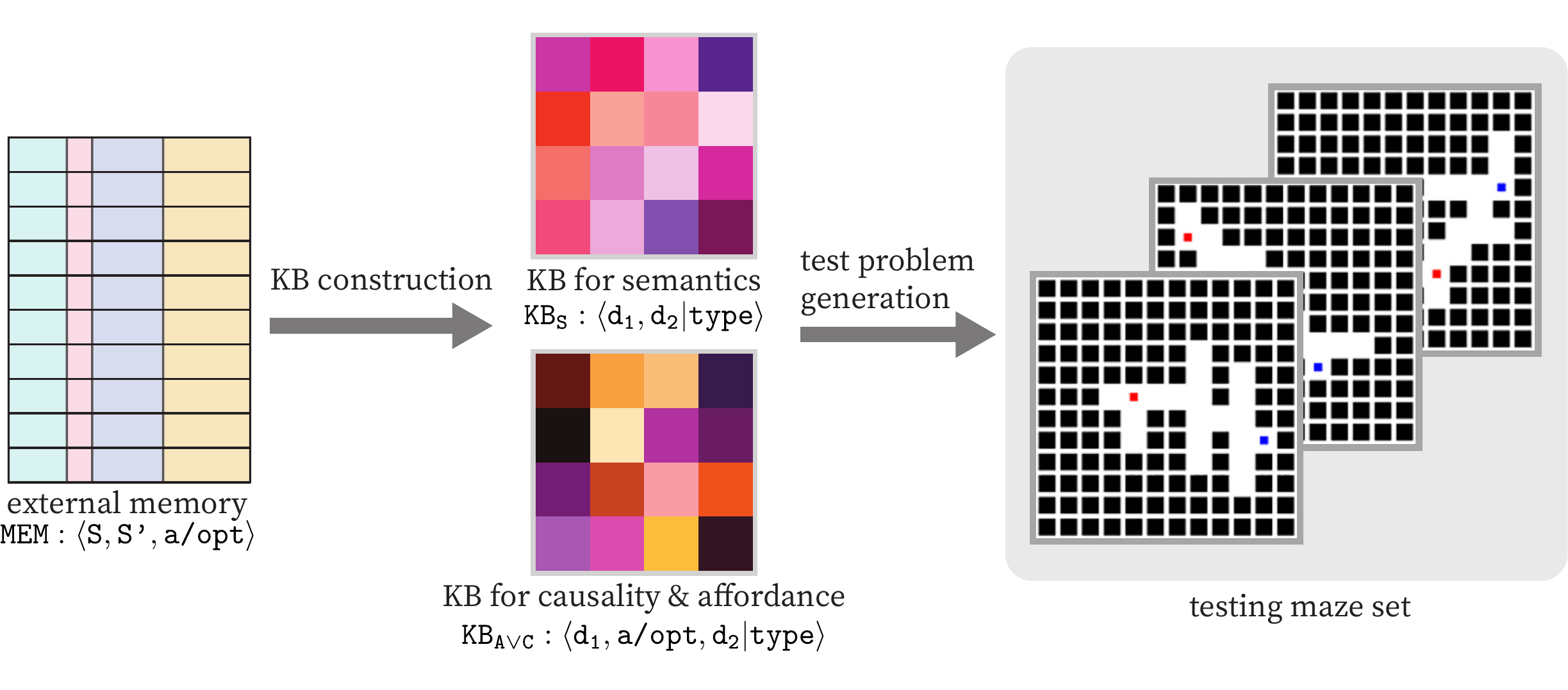}
    \caption{Illustration of the testing maze generation pipeline.}
    \label{fig:appx:testing_maze_gen}
\end{figure}

One of the unique features that \ac{halma} possesses is its capability of pinpointing the model weaknesses by dynamically generating informative and definitive generalization tests according to agents' experience. During training, we save the running experience of the agent as its external memory $\texttt{MEM}$, specifically as a tuple, containing (i) a pair of states $\texttt{s}$ and $\texttt{s'}$, and (ii) the action/option $\texttt{a/opt}$ the agent takes in this transition. Based on this external memory $\texttt{MEM}$, we build a pipeline that automatically generates the diagnostic testing set that tests a range of generalization abilities.

\paragraph{Knowledge Base Construction}
As shown in \cref{fig:appx:testing_maze_gen}, we first construct the Knowledge Base (KB) from the external memory $\texttt{MEM}$ by converting the tuples to inequality hitmaps following these rules:
\begin{itemize}[leftmargin=*,noitemsep,nolistsep]
    \item If a pair of \texttt{red}, \texttt{orange}, \texttt{yellow}, or \texttt{green} digits $\langle \texttt{d}_\texttt{1}, \texttt{d}_\texttt{2} \rangle$ occur in the same panel, then they are considered to represent the relation $\langle \texttt{d}_\texttt{1}, \texttt{d}_\texttt{2} | \texttt{type} \rangle$ that belongs to semantics inequality, where $\texttt{d}_\texttt{1}$ is the greater digit and $\texttt{d}_\texttt{2}$ the lesser one. Recall that the color mentioned above are connected with directions $\caretu$, $\caretd$, $\caretl$, and $\caretr$, respectively. In short, this KB is for $\langle\texttt{S}, <\rangle$. 
    \item If the digit colored \texttt{red}, \texttt{orange}, \texttt{yellow}, or \texttt{green} changes in states $\texttt{s}$ and $\texttt{s'}$, and if both digits are non-zero, then they are considered to represent the relation $\langle \texttt{d}_\texttt{1}, \texttt{d}_\texttt{2} | \texttt{type} \rangle$ that belongs to affordance inequality and causality equality, where $\texttt{d}_\texttt{1}$ is the greater digit in $\texttt{s}$ and $\texttt{s'}$ and $\texttt{d}_\texttt{2}$ the lesser one. In short, this KB is for $\langle\texttt{A},+/-, <\rangle\cup\langle\texttt{C},+/-, =\rangle$. 
    \item If the digit colored \texttt{red}, \texttt{orange}, \texttt{yellow}, or \texttt{green} appears in state $\texttt{s}$ and disappear in $\texttt{s'}$, meaning that the agent consumes the distance the digit $\texttt{d}$ represents, we consider the indication of that digit is revealed to the agent. We can therefore consider that the relation $\langle \texttt{d}_\texttt{1}, \texttt{d}_\texttt{2} | \texttt{type}\rangle$ that belongs to affordance inequality is explored, where $\texttt{d}_\texttt{1}$ and $\texttt{d}_\texttt{2}$ are digits understood through affordance, and $\texttt{d}_\texttt{1}$ is the greater digit and $\texttt{d}_\texttt{2}$ the lesser one. In short, this KB is for $\langle\texttt{A},+/-, =\rangle$.
\end{itemize}
% (Add compositional inequality & m-hop)

\paragraph{Test Problem Generation}
We pull inequality pairs from the constructed KB according to \cref{sec:gen_test} to generate the testing mazes. Specifically, each inequality relation pair $\langle \texttt{d}_\texttt{1}, \texttt{d}_\texttt{2} | \texttt{type}\rangle$ contains a pair of digits $\langle \texttt{d}_\texttt{1}, \texttt{d}_\texttt{2} \rangle$; we use the greater one as the distance till a wall, and the lesser one as the distance till a crossing. Therefore, we are able to incorporate the concepts to the maze layout and use the generated maze set to test the agent abilities of generalization. 

As mentioned in \cref{sec:gen_test}, generalization test problems in \ac{halma} are categorized into 3 different groups, \ie, \underline{S}emantic \underline{T}est (\textbf{ST}), \underline{Af}fordance \underline{T}est (\textbf{AfT}), and \underline{An}alogy \underline{T}est (\textbf{AnT}); there are also some more specific tests within each of these groups. Below, we provide detailed, concrete, and illustrative examples for each test unit:
\begin{itemize}[leftmargin=*,noitemsep,nolistsep]
    \item \textbf{ST}-1: We would like to know whether the agent can understand novel MNIST-digit-level combinations and make right decisions from those observations. Therefore, we pull inequality relation pairs that rarely co-occur in the external memory $\texttt{MEM}$ and create ST-1 mazes based on these digit pairs. For instance, if the agent observed a combination $\langle\mnist{3}{1}, \mnist{5}{1}, \mnist{1}{0}, \mnist{7}{5}\rangle$ during training, we would like to test whether the agent can make right decisions given $\langle\mnist{3}{1}, \mnist{5}{1}, \mnist{4}{3}, \mnist{2}{6}\rangle$, which is rarely or never observed during training, \ie, $\langle\mnist{3}{1}, \mnist{5}{1}, \mnist{4}{3}, \mnist{2}{6}\rangle\not\in\texttt{MEM}$. To achieve this goal, we can create a maze segment, where there is a crossing 3-grid away upwards and a wall 5-grid away in the same direction to ensure that the visual panel includes $\langle\mnist{3}{1}, \mnist{5}{1}\rangle$. We then add a wall 4-grid away downwards to include the $\mnist{4}{3}$, and set the goal state position 2-grid away downwards to include the $\mnist{2}{6}$. Finally, we can assemble this kind of maze segments to create the desired testing mazes.
    \item \textbf{ST}-2: We would like to know whether the agent can recognize novel digit attributes combinations. We focus on the novel combination of $\texttt{color}$ and MNIST $\texttt{category}$ in \textbf{ST}-2 mazes. For instance, if the agent observes $\langle\mnist{3}{1}, \mnist{5}{1}\rangle$ during training, we would test whether the agent can take right actions given $\langle\mnist{3}{3}, \mnist{5}{3}\rangle$, which should be never seen during training, \ie, $\langle\mnist{3}{3}, \mnist{5}{3}\rangle\not\in\texttt{MEM}$. We can create a maze segment, where there is a crossing 3-grid away downwards and a wall 5-grid away in the same direction to ensure that the visual panel includes $\langle\mnist{3}{3}, \mnist{5}{3}\rangle$.
    \item \textbf{Aft}-1: We would like to know whether the agent can understand the indication of MNIST digits through causal transitions in \textbf{Aft}-1. For instance, if the agent observed the $\mnist{5}{0}$ in the visual panel, moved 2 steps to the left $\caretl :\dicetwo$, and observed the $\mnist{3}{0}$, we would expect the agent to understand that $\mnist{3}{0} < \mnist{5}{0}$ through this transition. We can directly pull this kind of inequality pairs from $\texttt{KB}_{\texttt{A}\lor\texttt{C}}$. Note that to create pure testing mazes for \textbf{Aft}-1, we need to ensure that there are neither direct observations of the digit pair $\langle\mnist{3}{0},\mnist{5}{0}\rangle$ from visual panels, nor visual observations of $\langle\mnist{3}{1},\mnist{5}{1}\rangle$, $\langle\mnist{3}{2},\mnist{5}{2}\rangle$, or $\langle\mnist{3}{3},\mnist{5}{3}\rangle$, which would help to infer the inequality relation $\mnist{3}{0} < \mnist{5}{0}$ if the agent could recognize novel digit attributes combinations as in \textbf{ST}. In short, if an inequality pair $\langle\texttt{d}_\texttt{1},\texttt{d}_\texttt{2}\rangle$ is pulled from $\texttt{KB}_{\texttt{A}\lor\texttt{C}}$ to create the testing mazes for $\textbf{Aft}$-1, then we must have $\langle\texttt{d}_\texttt{1},\texttt{d}_\texttt{2}\rangle\not\in\texttt{KB}_\texttt{S}$. We can then similarly create a maze segment as in \textbf{ST}.
    \item \textbf{Aft}-2: We would like to know whether the agent can understand the indication of MNIST digits through affordance in \textbf{Aft}-2. For instance, if the agent exploited the affordance of $\mnist{5}{1}$ with $\caretu : \dicetwo+\dicethree$ and exploited the affordance of $\mnist{3}{1}$ with $\caretu : \dicethree$ during training, we would expect it to understand these two digits. We can directly pull such inequality pairs from $\texttt{KB}_\texttt{A}$ to create testing mazes for \textbf{Aft}-2. Note that the inequality pair $\langle\texttt{d}_\texttt{1},\texttt{d}_\texttt{2}\rangle$ pulled from $\texttt{KB}_\texttt{A}$ should not be in $\texttt{KB}_\texttt{S}$ for the same reason as in \textbf{Aft}-1.
    \item \textbf{Aft}-3 and \textbf{Aft}-4: We would like to know whether the agent can understand the indication of MNIST digits through transitions and affordance based on their understanding of the composition of visual attributes. For instance, if the agent's causality and affordance knowledge base $\texttt{KB}_{\texttt{A}\lor\texttt{C}}$ included the inequality pair $\langle\mnist{3}{1},\mnist{5}{1}\rangle$ or $\langle\mnist{3}{3},\mnist{5}{2}\rangle$, we would test whether the agent can understand $\langle\mnist{3}{0},\mnist{5}{0}\rangle$. Note that we need to ensure that $\langle\mnist{3}{0},\mnist{5}{0}\rangle$ is not in the $\texttt{KB}_\texttt{S}$.
    \item \textbf{AnT}: Note that in \textbf{ST} and \textbf{Aft}, we only test the direct inequality relation between digits. Here, we test the agent's understanding of transitivity of inequality relations in \textbf{AnT}. We expect agents to acquire the understanding of transitivity with analogical reasoning. For instance, if the agent's $\texttt{KB}_{\texttt{AfT}\lor\texttt{ST}}$ included a analogical template $\{\langle\mnist{3}{0},\mnist{4}{0}\rangle, \langle\mnist{4}{1},\mnist{5}{1}\rangle, \langle\mnist{3}{2},\mnist{5}{2}\rangle\}$, we would expect agents to learn analogical reasoning from this base case. If there was another pair of tuples $\langle\mnist{6}{3}, \mnist{7}{3}\rangle$, $\langle\mnist{7}{1}, \mnist{8}{1}\rangle$ in $\texttt{KB}_{\texttt{AfT}\lor\texttt{ST}}$, and further given that $\langle\mnist{6}{0},\mnist{8}{0}\rangle$ was not in the $\texttt{KB}_{\texttt{AfT}\lor\texttt{ST}}$, we would test the agent's understanding of transitivity from the analogical template.
\end{itemize}
% It is worth mentioning that these test problems are mutually exclusive.
Recall that in HALMA, we use 10 MNIST categories to indicate the distance till a wall or the nearest crossing, from which we extract the inequality relations and form the knowledge base. The number of inequality pairs is thus limited. Because the test units listed above are mutually exclusive, it is likely that some of the test problems may not be generated if the agent's experience, along with already generated tests, cover the full space of inequality. This explains the ``-'' in \cref{tab:exp_results}.

\clearpage

\section{Details of Models}
\label{appx:models}

\subsection{Hyper-parameters of TD3}

\begin{table}[ht!]
    \centering
    \caption{Hyper-parameters of TD3}
    \label{tab:appx:hyper_param}
    \begin{tabular}{lc}
        \toprule
        Hyper-parameters                            &Value \\ \hline
        Optimizer                                   &Adam \citep{kingma2014adam} \\
        Learning rate for actor                     &1e-4 \\
        Batch size                                  &128 \\
        $\epsilon$ of Adam                          &1e-8 \\
        % Number of critics in ensemble               &2 \\
        Discounting factor                          &0.95 \\
        Initial $\epsilon$ for $\epsilon$-greedy    &0.1 \\
        Ending $\epsilon$ for $\epsilon$-greedy     &0.95 \\
        Decay steps for $\epsilon$-greedy           &100,000 \\
        Policy update delay                         &5 \\
        Target update rate                          &0.995 \\
        Replay buffer size                          &10,000 \\
        \bottomrule
    \end{tabular}
\end{table}

\subsection{Architecture of Agents}
\label{sec:arch_agents}

\begin{figure}[th!]
    \centering
    \includegraphics[width=0.7\linewidth]{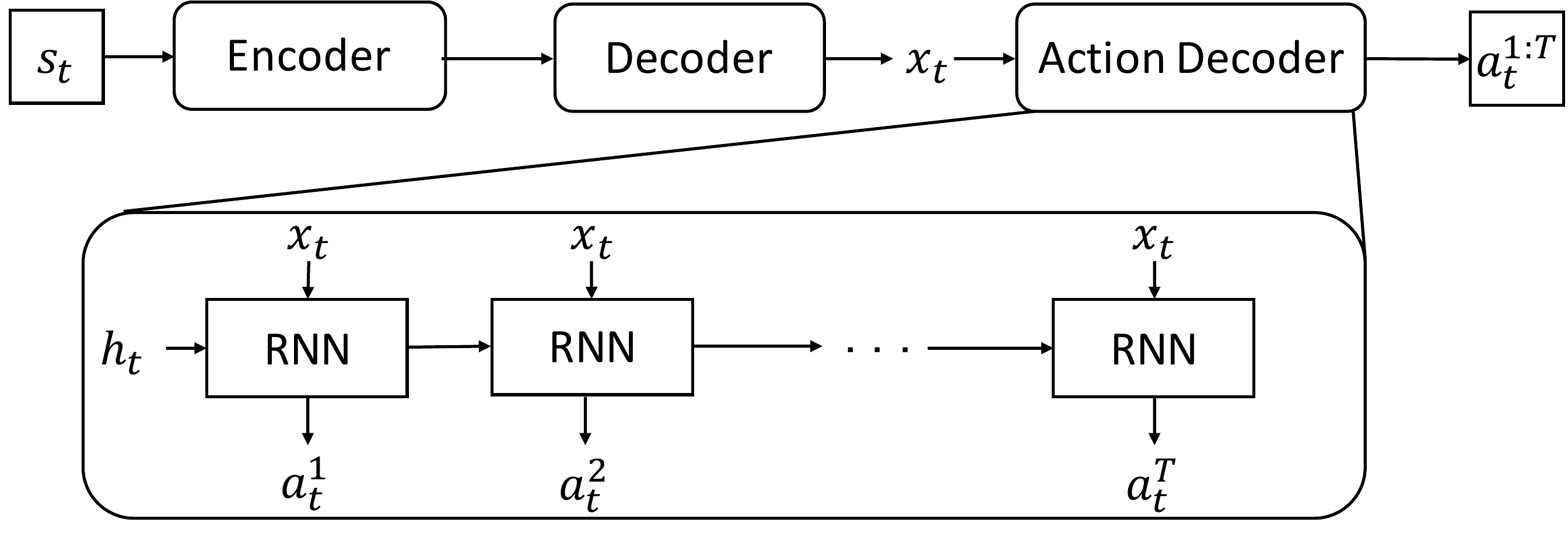}
    \caption{Architecture of the actor model, where T is equal to \texttt{max\_opt\_len}.}
    \label{fig:actor_arch}
\end{figure}

\begin{figure}[th!]
    \centering
    \includegraphics[width=0.7\linewidth]{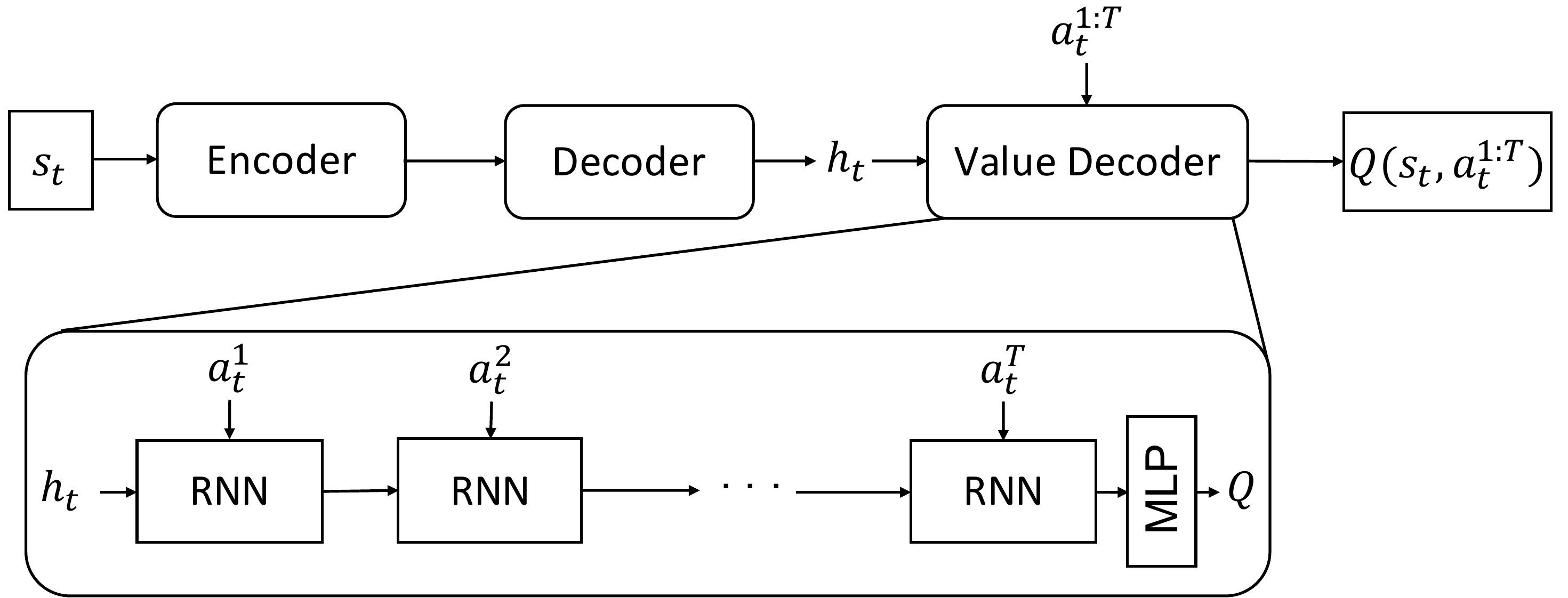}
    \caption{Architecture of the critic model, where T is equal to \texttt{max\_opt\_len}.}
    \label{fig:critic_arch}
\end{figure}

The overall architectures of the actor and the critic model employed by our agents are illustrated in \cref{fig:actor_arch} and \cref{fig:critic_arch}, respectively. All agents share the same implementation of the action decoder and the value decoder, which allows them to work with action sequences, \ie, option. Note that the hidden vector $h_t$ in the actor is simply initialized as a zero vector, and the critic uses the output of decoder instead to condition its output Q value on the state input.

The major difference among agents lies in the implementation of their inductive biases, \ie, encoder and decoder. We provide a summary, along with some other hyperparemeters, in \cref{tab:appx:shared_params}.

\begin{table}[ht!]
    \centering
    \caption{Architectural parameters of evaluated agents} 
    \label{tab:appx:shared_params}
    \begin{tabular}{l l| l }
        \toprule
        \multicolumn{2}{l|}{Agent} & Architecture\\
        \midrule
        \multicolumn{2}{l|}{\it{Shared}}& \\
        & Nonlinearity & ReLU\\
        \midrule
        \multicolumn{2}{l|}{\it{MLP Agent}}& \\
        & Encoder & \makecell[tl]{MLP with hidden units [128, 128].} \\
        & Decoder & None\\
        \midrule
        \multicolumn{2}{l|}{\it{LSTM Agent}}& \\
        & Encoder & \makecell[tl]{MLP with hidden units [128, 128].} \\
        & Decoder & \makecell[tl]{LSTM with layer normalization \citep{ba2016layer} and\\
        hidden units [128].}\\
        \midrule
        \multicolumn{2}{l|}{\it{Transformer Agent}}& \\
        & Encoder & \makecell[tl]{A stack of four multi-head self-attention layers, with hidden \\
        units [128], four heads and layer normalization, followed by a \\
        maximum pooing layer. Parameters are shared across all the \\
        attention layers \citep{zambaldi2018deep}.} \\
        & Decoder & MLP with hidden units [128, 128]. \\
        \midrule
        \multicolumn{2}{l|}{\it{Transformer+LSTM Agent}}& \\
        & Encoder & \makecell[tl]{Identical to \it{Transformer Agent}.} \\
        & Decoder & \makecell[tl]{MLP with hidden units [128, 128], followed by LSTM with layer \\
        normalization \citep{ba2016layer} and hidden units [128].}\\
        \midrule
        \multicolumn{2}{l|}{\it{CNN Agent}}& \\
        & Encoder & \makecell[tl]{CNN with kernel parameters [(3, 32, 6, 4), (32, 64, 6, 4), \\
        (64, 128, 7, 1)] (number of input filters, number of output filters,\\
        kernel size, and stride size by ordering).} \\
        & Decoder & \makecell[tl]{MLP with hidden units [128].}\\
        \midrule
        \multicolumn{2}{l|}{\it{CNN+Transformer Agent}}& \\
        & Encoder & \makecell[tl]{CNN with kernel parameters [(3, 32, 4, 4, 0), (32, 64, 4, 4, 0), \\
        (64, 128, 3, 2, 1)] (number of input filters, number of output filters,\\
        kernel size, stride size, and padding size by ordering); resized \\
        to $4\times4$ slots, concatenated with positional embedding\\
        (\cref{appx:exp_protocol}); followed by the encoder of \it{Transformer Agent}.} \\
        & Decoder & \makecell[tl]{Identical to \it{Transformer Agent}.}\\
        \midrule
        \multicolumn{2}{l|}{\it{SPACE Agent}}& \\
        & Encoder & \makecell[tl]{We adopt the original setup of SPACE \citep{lin2019space} for the\\ 
        \texttt{image\_encoder} and the \texttt{what\_encoder}. We concatenate latent \\
        vectors for the shape ($Z_{what}$) and the presence ($Z_{where}$) of \\
        each object. In sum, there are $8\times8$ object slots, each is a 33-D \\
        vector. They are then fed to the encoder of \emph{Transformer Agent}.} \\
        & Decoder & \makecell[tl]{Identical to \it{Transformer Agent}.}\\
        \bottomrule
    \end{tabular}
\end{table}

\clearpage
\setstretch{0.99}

\section{Experimental Details}
\label{appx:exp}

\subsection{Task Parameters and Experimental Protocol} 
\label{appx:exp_protocol}

\paragraph{Task parameters}
Task parameters for \ac{halma} are mostly defined in \cref{sec:task_formulation}, explicitly specified for the formulation of rapid problem solving. \cref{tab:appx:task_param} summarizes these parameters.

\begin{table}[ht!]
    \centering
    \caption{Task parameters of \ac{halma}}
    \label{tab:appx:task_param}
    \begin{tabular}{lc}
        \toprule
        Task parameters                         & Value \\ \hline
        Maximum \#steps in an episode ($L$)     &500 \\
        Maximum \#trials in an episode ($N$)    &10 \\
        Maximum \#steps in a trial($H$)         &200 \\
        Discounting factor ($\gamma$)           &0.95 \\
        Goal reward ($R_g$)                     &100 \\
        Penalty on invalid action ($R_a$)       &-5 \\
        Penalty on invalid action ($R_a$)       &-5 \\
        Auxiliary rewards ($R_x$)               &\makecell[tl]{1.0$\times (L_m(\text{agent}_{t-1},\text{goal})$ \\
        $-L_m(\text{agent}_{t},\text{goal}))$, where \\
        $L_m$ is the Manhattan distance.} \\
        \bottomrule
    \end{tabular}
\end{table}

Given a set of generated \ac{halma} problems, there is still one task parameter: $\texttt{max\_opt\_len}$, which is the maximum length of an option in one step. We tried three different setups, $\{1,3,5\}$. Intuitively, when $\texttt{max\_opt\_len=1}$, agents do not need to merge sub-options to improve planning efficiency, though they may still need to decide between $\{\dicez,\diceone,\dicetwo,\dicethree\}$. With that said, the exploration and planning efficiency $\bm{\rho_p}$ may be close to the optimal $1$ as long as the ratio of goal reaching $\bm{\rho_g}$ is high. In contrast, when $\texttt{max\_opt\_len}=3\text{ or }5$, agents would need to understand the compositionality of the option space (\ie, the temporal grammar) to improve $\bm{\rho_p}$. In this case, as shown in \cref{appx:training_curves}, most agents find it quite challenging to \emph{plan optimally}. They may even get trouble in understanding affordance, hence have a lower ratio of valid moves $\bm{\rho_a}$ than when $\texttt{max\_opt\_len=1}$. 

\paragraph{Two types of observations}
We provide two types of observations to the agents. One is a low-dimensional \emph{symbolic observation space}. It represents the ground-truth MNIST digits, colors, and shape of hint symbols at crossing. Recall that in \ac{halma}, the observation may have at most 10 MNIST digits\footnote{4 for crossings, 4 for distance to the walls, and the remainder, 2, for distance to the goal.} plus 1 crossing hint, and the value of digit range from -9 to 9,\footnote{For the two goal digits only, while others are only allowed to be in $\{\texttt{0},\texttt{1},\texttt{2},\texttt{3},\texttt{4},\texttt{5},\texttt{6},\texttt{7},\texttt{8},\texttt{9}\}$; the hint only has 4 different values)} which results in 10 one-hot vectors with an overall size of $11\times19$. For agents with permutation invariant modules (\eg, transformers), we enforce the positional sensitivity by augmenting each one-hot vector with an extra indexing vector of size 10, which is essentially another one-hot vector that indicates the index. In our experiments, we observe that this index encoding is crucial to all the transformer-based agents.

We also offer a \emph{visual observation space}, where the only observation is the visual panel of \ac{halma}, as introduced in \cref{sec:halma_basics,sec:problem_generation}. We downsample them to a RGB image with size $(128, 128, 3)$ and re-scaled to $[0, 1]$. Agents for this type of observations require visual modules, such as CNN or SPACE \citep{lin2019space} as detailed in \cref{sec:arch_agents}.

\paragraph{Training protocol}
We generated 100 mazes for training. An ablation study on the volume of training set can be found in \cref{appx:ablation}. Each agent is trained for 2000 episodes under the task formulation introduced in \cref{sec:task_formulation}. All of them converged at the end of training, as illustrated in their learning curves in \cref{appx:training_curves}. We tried 5 different seeds during training and report the best result. Note that different from classical reinforcement learning tasks, where there is no explicit split for training and testing hence training curves are reported for quantitative evaluation, we provide training curves merely for justifying the validity of our training. 

\paragraph{Testing protocol}
We test all agents in (i) the training problems, (ii) test problems generated by random split in the problem space, and (iii) test problems dynamically generated according to \cref{sec:gen_test}. The former two are provided mainly for reference. Interestingly, most agents perform almost equally well on these two, consistent with prior works \citep{guez2019investigation,cobbe2019quantifying}. For all tests or dynamically generated subtests, we test with \~150 mazes and summarize over 3 different seeds to calculate mean and standard deviation. A test is skipped if the dynamic generation fails, as introduced in \cref{appx:kb}.

\setstretch{1}

\subsection{Learning Curves}
\label{appx:training_curves}

\begin{figure}[th!]
    \centering
    \includegraphics[width=\linewidth]{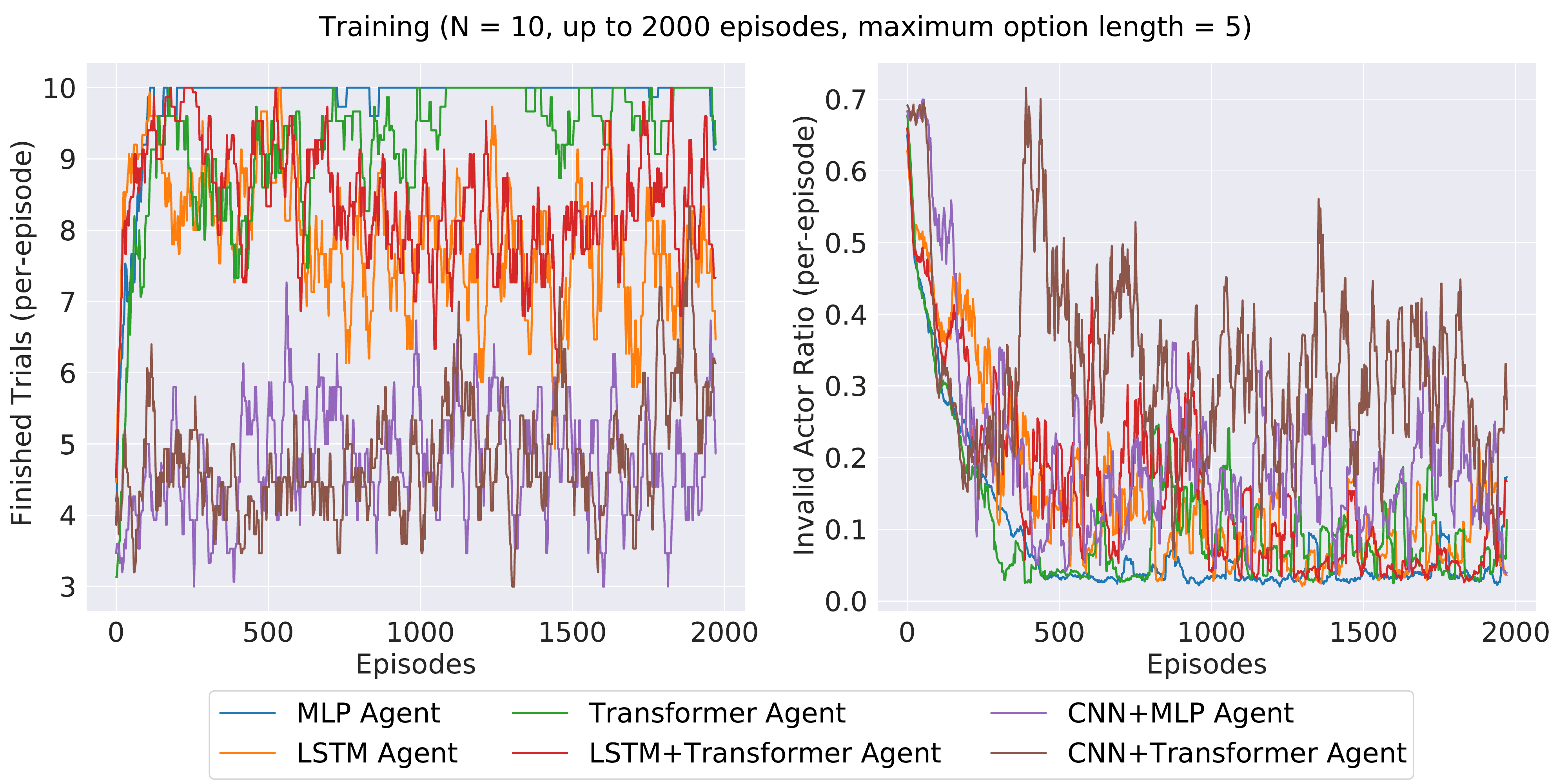}
    \caption{Learning curves of the evaluated agents with \texttt{max\_opt\_len=5}.}
    \label{fig:appx:train_curve_alen5}
\end{figure}

\begin{figure}[th!]
    \centering
    \includegraphics[width=\linewidth]{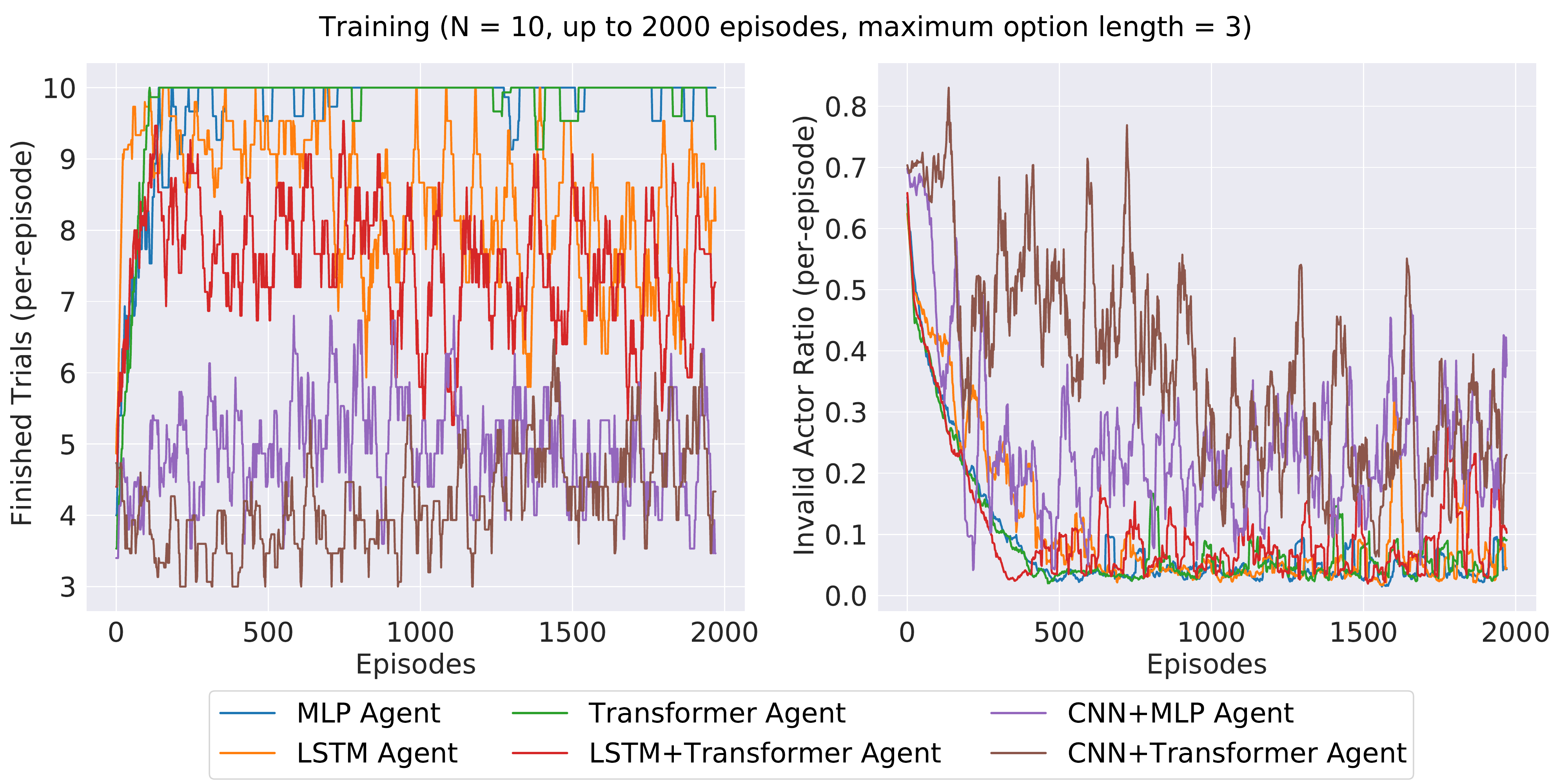}
    \caption{Learning curves of the evaluated agents with \texttt{max\_opt\_len=3}.}
    \label{fig:appx:train_curve_alen3}
\end{figure}

\begin{figure}[th!]
    \centering
    \includegraphics[width=\linewidth]{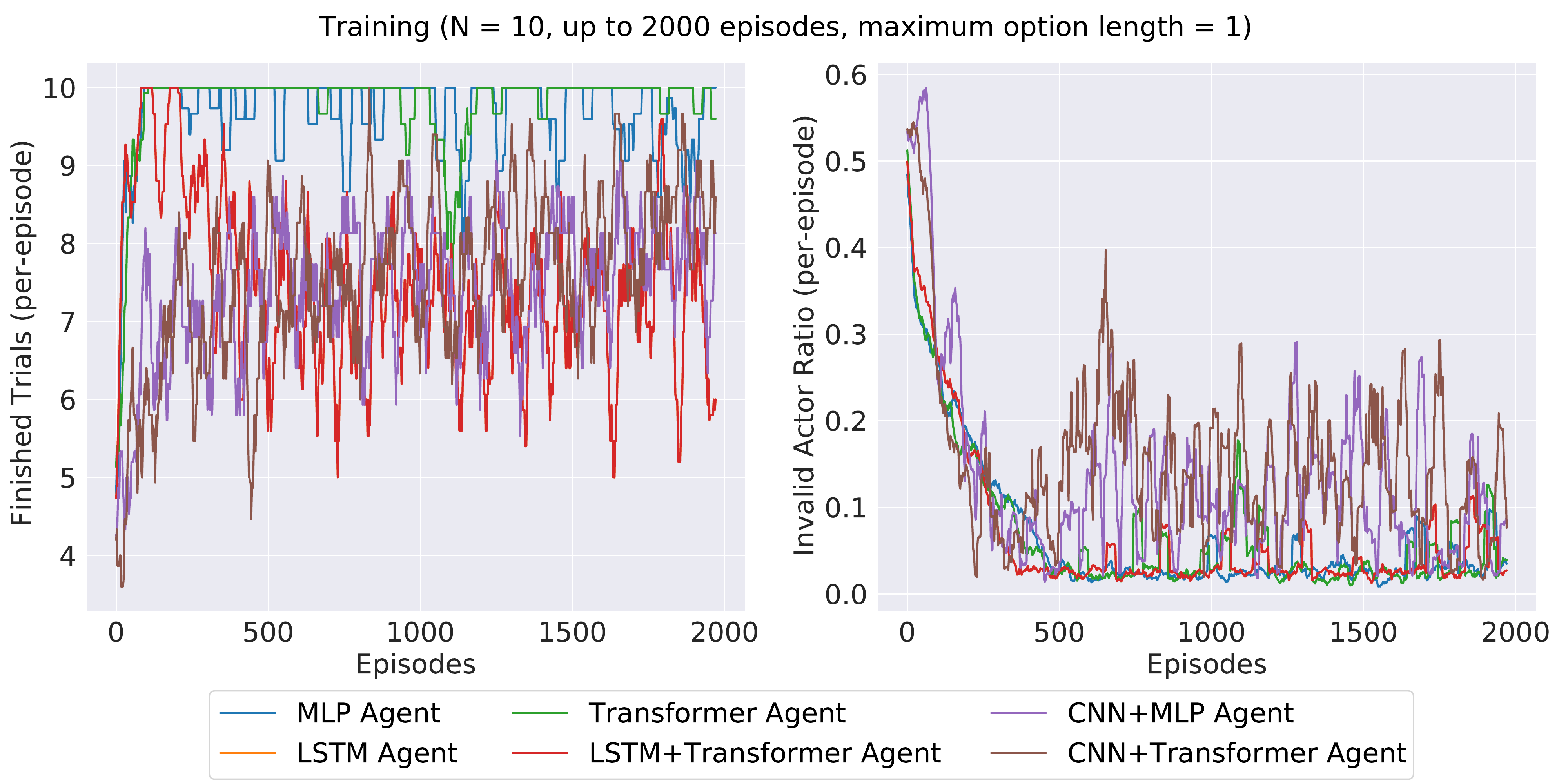}
    \caption{Learning curves of the evaluated agents with \texttt{max\_opt\_len=1}.}
    \label{fig:appx:train_curve_alen1}
\end{figure}

To validate the convergence during training, we provide the learning curves of agents trained under different settings (mainly on the different choice of $\texttt{max\_opt\_len}$) in \cref{fig:appx:train_curve_alen5,fig:appx:train_curve_alen3,fig:appx:train_curve_alen1}. We report the number of finished trials and the ratio of valid moves in each training episode. The moving average (with a window size of the number of mazes in the training set) of these two metrics can reflect $\bm{\rho_g}$ and $\bm{\rho_a}$ in training. These curves suggest that all agents with symbolic observations converge before 2000 episodes in terms of the goal reaching rate and valid moves ratio. For the visual observation, however, agents struggles on both metrics when the action space is large ($\texttt{max\_opt\_len=3}$ or $\texttt{max\_opt\_len=5}$). Their performances remain almost the same after 2000 episodes. Hence, we report the test results with $\texttt{max\_opt\_len=1}$ in the main paper; full results can be found below. 

\subsection{SPACE Model}
\label{appx:space}

\paragraph{Architecture and Hyperparemeters}
We adopt the original setup of SPACE \citep{lin2019space} except for a simple modification in the background encoder. Specifically, we replace their \texttt{StrongCompDecoder} with their \texttt{CompDecoder}.

\begin{figure}[th!]
    \centering
    \begin{subfigure}[b]{0.25\linewidth}
        \centering
        \includegraphics[width=\linewidth]{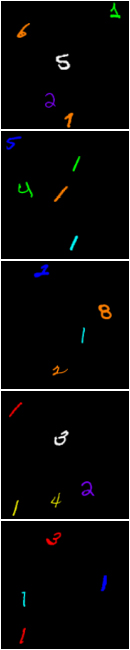}
        \caption{Input}
    \end{subfigure}%
    \begin{subfigure}[b]{0.25\linewidth}
        \centering
        \includegraphics[width=\linewidth]{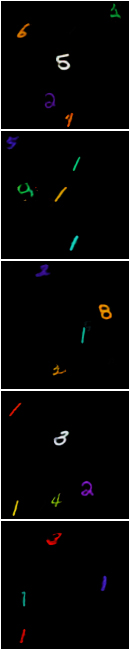}
        \caption{Reconstruction}
    \end{subfigure}%
    \begin{subfigure}[b]{0.25\linewidth}
        \centering
        \includegraphics[width=\linewidth]{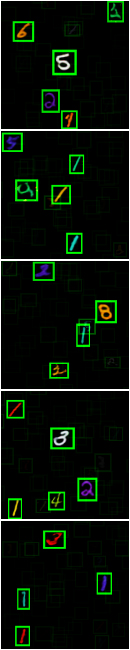}
        \caption{Detection}
    \end{subfigure}%
    \begin{subfigure}[b]{0.25\linewidth}
        \centering
        \includegraphics[width=\linewidth]{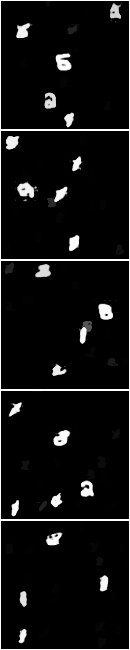}
        \caption{Segmentation}
    \end{subfigure}%
    \caption{Visualization of SPACE's reconstruction, detection, and segmentation on hold-out \textbf{testing set}.}
    \label{fig:appx:space_vis}
\end{figure}

\paragraph{Reconstruction, Segmentation and Detection on Testing Set}
We train the SPACE model with all visual panels in the training set. To qualitatively evaluate the generalization capability of the SPACE model, we visualize their inference results in a hold-out testing set; it is essentially a set of visual panels from randomly generated test problems; see \cref{fig:appx:space_vis} for an example. The SPACE model generalizes remarkably well in terms of reconstruction, detection, and segmentation, consistent with the original results reported by \citep{lin2019space}. 

\setstretch{1}

\paragraph{Investigating the Latent Space}
% Intro
We further investigate the efficacy of the SPACE model in disentangling independent latent factors from visual panels. Specifically, we adopt a standard methodology in the unsupervised disentanglement learning literature \citep{higgins2016beta}, linear probing.

% Setup & Implementation
We train a linear SVM classifier using the latent representations of colored MNIST digits obtained from the encoder of the SPACE model. We observe that the output vector of SPACE encoder have multiple slots representing the objects (digits) in the input image, and that the connection between slots and input objects is implicit. Hence, we calculate the IoU of predicted bounding box and ground-truth bounding box to assign each slot to an input object as its semantic label. 
In this work, there are 64 slots in the output vector and no more than 11 objects in the input image. Therefore, it is likely that several slots are assigned to the same object. We save for each object only the slot with the maximum IoU to remove redundancy in the data and obtain $8,932$ 33-D latent vectors in total. 
We use 70\% of these samples as training data and perform testing on the held-out 30\% data by randomly splitting the latent vectors. We set the penalty parameter `C' of SVM as 10 in all experiments and use balanced sampling when training the classifier. SVM classifier is implemented with the scikit-learn package \citep{scikit-learn}.

\begin{figure}[th!]
    \centering
    \begin{subfigure}[b]{0.48\linewidth}
        \centering
        \includegraphics[width=\linewidth]{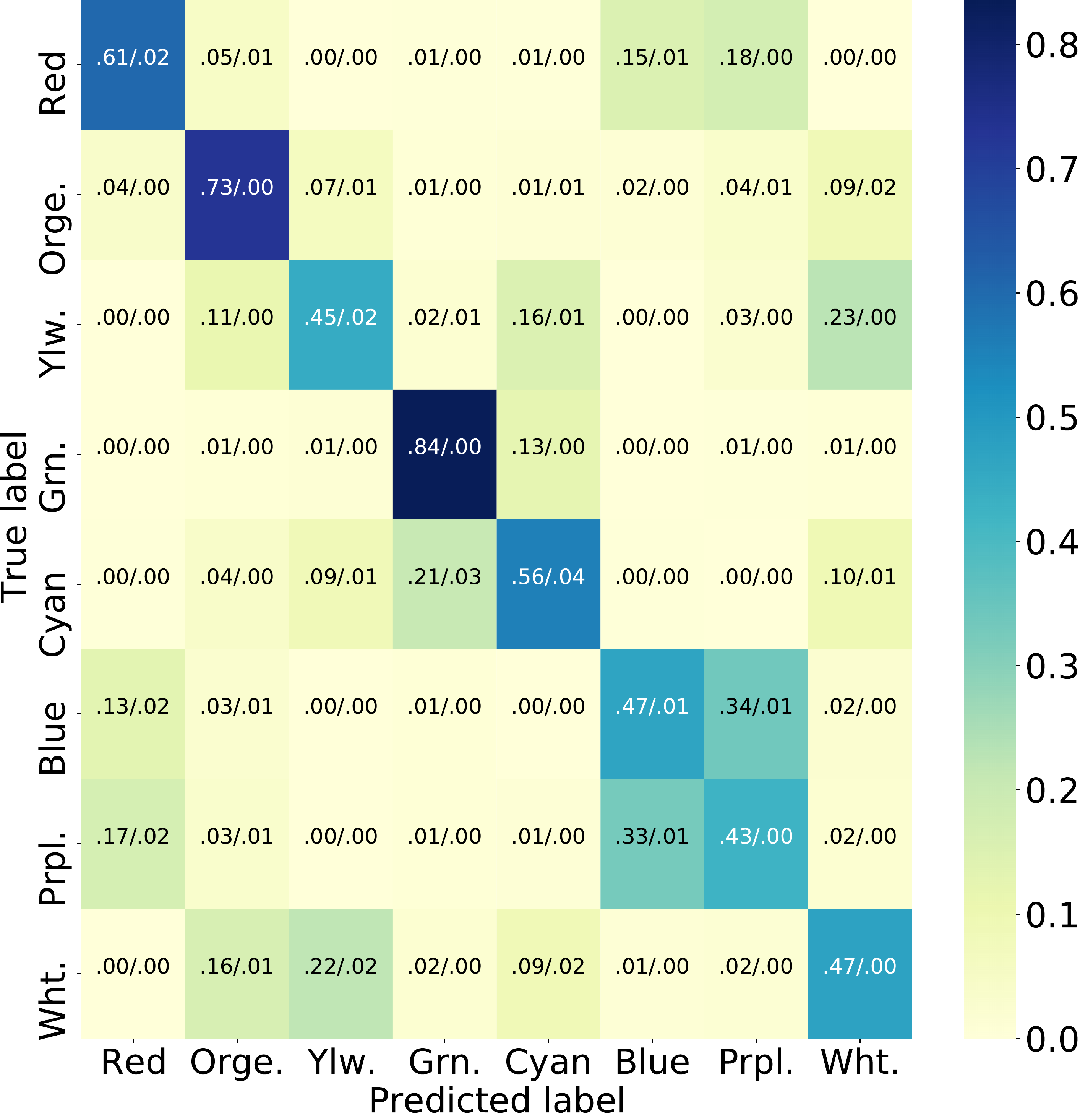}
        \caption{}
    \end{subfigure}%
    \hfill%
    \begin{subfigure}[b]{0.48\linewidth}
        \centering
        \includegraphics[width=\linewidth]{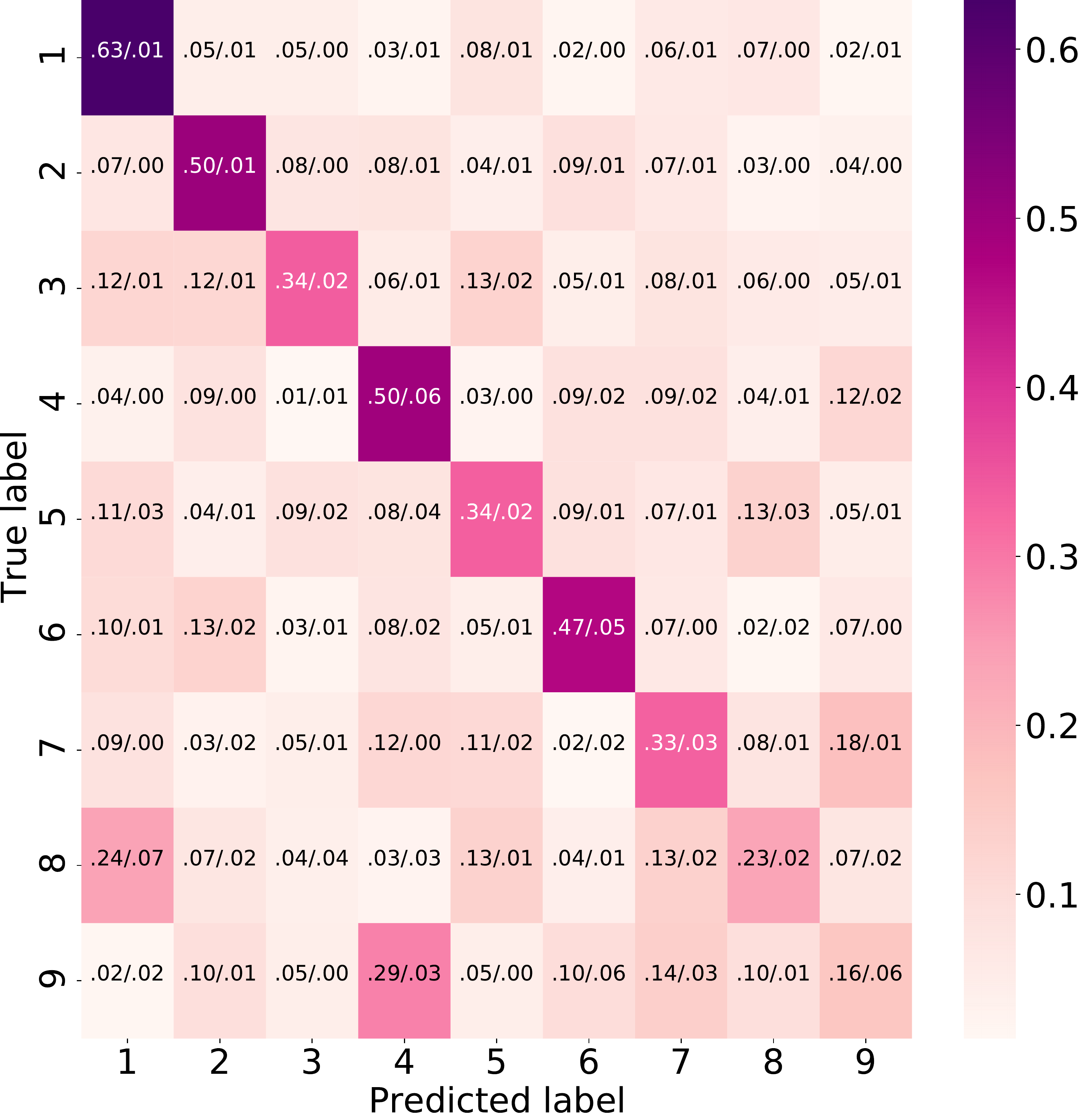}
        \caption{}
    \end{subfigure}%
    \caption{Confusion matrix of linear classifier trained on latent representations from SPACE encoder. Confusion matrix of (a) \texttt{color} and (b) MNIST \texttt{category} are measured and averaged over 10 random split testing set. Mean value and standard deviation of the accuracy are displayed in the matrix.}
    \label{fig:conf_mat}
\end{figure}

\begin{table}[ht!]
    \centering
    \caption{Accuracy of \texttt{color} and MNIST \texttt{category} classification.}
    \label{tab:SPACE_latent_classification}
    \begin{tabular}{l|ccccccccc}
        \toprule
        Task & \texttt{color} & MNIST \texttt{category} \\
        \midrule
        Acc. & \ditem{59.67}{0.58} & \ditem{50.56
}{0.92}\\
        \bottomrule
    \end{tabular}
\end{table}

% Result
We test the classification accuracy in terms of \texttt{color} and MNIST \texttt{category} and report the overall accuracy in \cref{tab:SPACE_latent_classification}. Each result is averaged over 10 random split of latent vectors. In addition, we provide the confusion matrix of these two attributes (\cref{fig:conf_mat}) to illustrate the categorical accuracy. Results in \cref{fig:conf_mat} (a) demonstrates that the SPACE model performs relatively well on the first four colors, \ie, \texttt{red}, \texttt{orange}, \texttt{yellow}, and \texttt{green}, while poorly on the rest. It partly explains SPACE agents' high invalid move ratio $\bm{\rho_a}$ and low goal reaching ratio $\bm{\rho_g}$ in \ac{halma}, \ie, agents cannot tell the correct direction. Results in \cref{fig:conf_mat} (b) demonstrates that the SPACE model does not handle the long-tail distribution of digits, and partly explains SPACE agents' high invalid move ratio $\bm{\rho_a}$ and low efficiency ratio $\bm{\rho_p}$ in \ac{halma}, \ie, agents do not know ``what it is'' in the first place.

\section{Additional Experiments}
\label{appx:ablation}

\subsection{Ablation Study on the Volume of Training Set}
\label{appx:ablation_trainset}

\begin{figure}[th!]
    \centering
    \includegraphics[width=\linewidth]{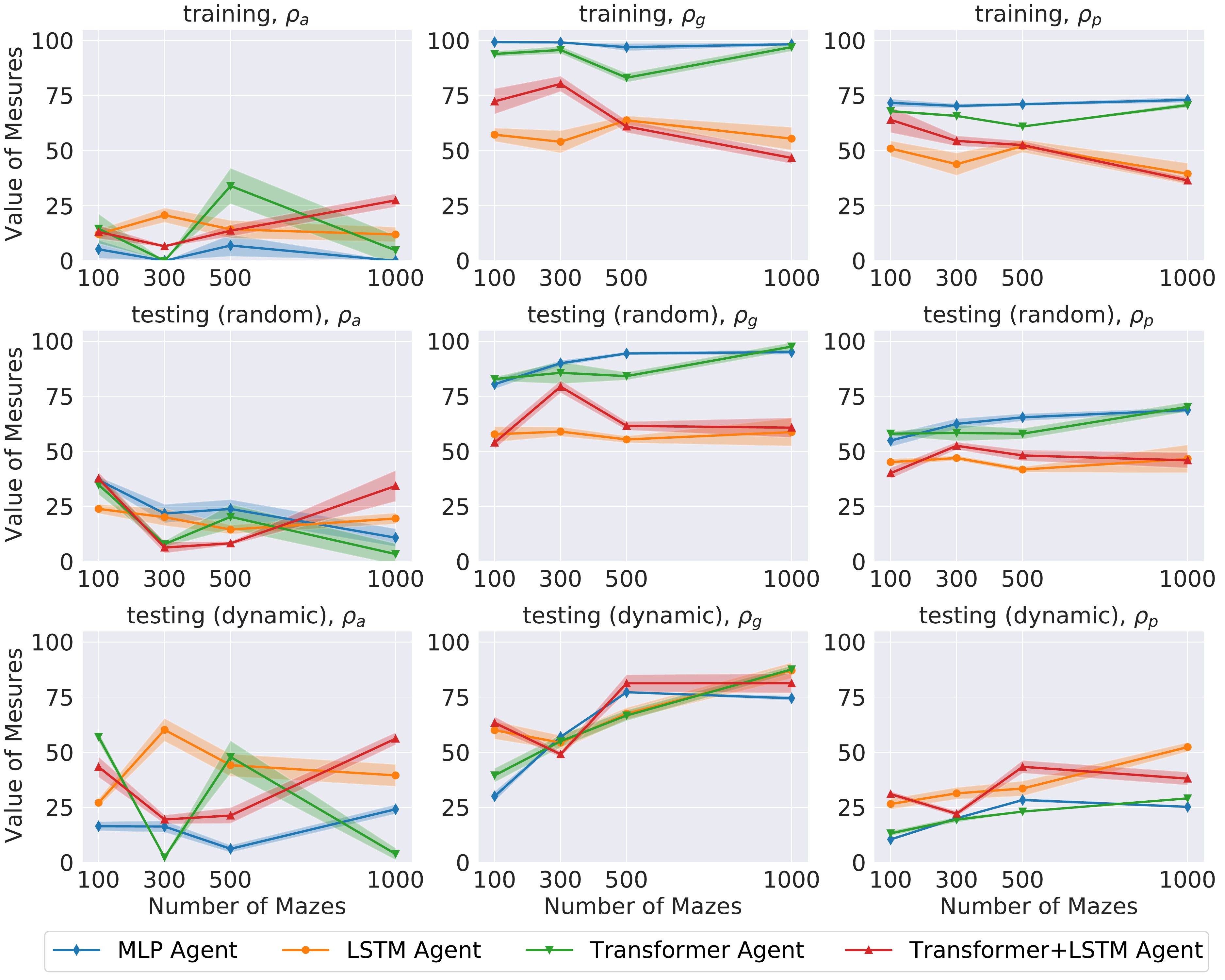}
    \caption{Ablation study of different number of training mazes.}
    \label{fig:appx:abla_maze}
\end{figure}

The thesis argument of our work is that humanlike agents shall generalize their understanding under limited exposure to the underlying concept spaces. To further investigate how the degree of exposure would affect agents performance in \ac{halma}, we first conduct an ablations study with different numbers of training mazes. Specifically, we experiment with four setups of the maze quantity for agents to explore during training: $100, 300, 500, 1000$ (results of 100 training mazes are reused from the main experiment as it is our default setting). Here we only evaluate agents with symbolic input: MLP agents, LSTM agents, Transformer agents and Transformer+LSTM agents. We report the three measures $\bm{\rho_a}$, $\bm{\rho_g}$ and $\bm{\rho_p}$ with all the testing protocols (training problems, problems from random split in the problem space and dynamically-generated testing problems) in \cref{fig:appx:abla_maze}. Note that measures in dynamically-generated tests are merged across subtests for better comparison.

The results read that, all agents could gain a performance boost with increased exposure during training. Specifically, there is a significant promotion for the metric of goal reaching rate $\bm{\rho_g}$ in the challenging dynamic testing (from $30\text{-}60\%$ to $80\%$). More interestingly, starting from 300 training mazes, the distinction between different inductive biases vanishes. While the efficiency ratio $\bm{\rho_p}$ could also benefit from increased exposure, it reaches only around $50\%$ at best. As for the ratio of valid moves $\bm{\rho_a}$, even though it reaches around $90\%$ in random split for stateless agent when trained with 1000 mazes, no clear trend can be detected in dynamic testing overall, which may suggest agents' limitation in understanding affordance with the temporal grammar or under the long-tail distribution of digits.

\subsection{Ablation Study on the Maximum Option Length \texorpdfstring{$\texttt{max\_opt\_len}$}{}}
\label{appx:ablation_alen}

Our design to include the notion of option challenges agents' understanding in the temporal grammar and the causal structure. To further illustrate the difficulty of this specific challenge, we also perform an ablation study on three setups of maximum option length $\texttt{max\_opt\_len}$. In general, agents' performance degrades on all metrics with $\texttt{max\_opt\_len}$ increases. In particular, the ratio of valid moves $\bm{\rho_a}$ decreases and the efficiency ratio $\bm{\rho_p}$ drops significantly since $\texttt{max\_opt\_len=3}$ in dynamic testing, suggesting that agents all have  hard time understanding either the temporal grammar or the causal structure of \ac{halma}. These results validate our argument that significant efforts are still in need for humanlike abstraction learning. Therefore, we choose to make the length of 5 as our default setting in the main paper so as to make \ac{halma} a more challenging territory.

\begin{figure}[th!]
    \centering
    \includegraphics[width=\linewidth]{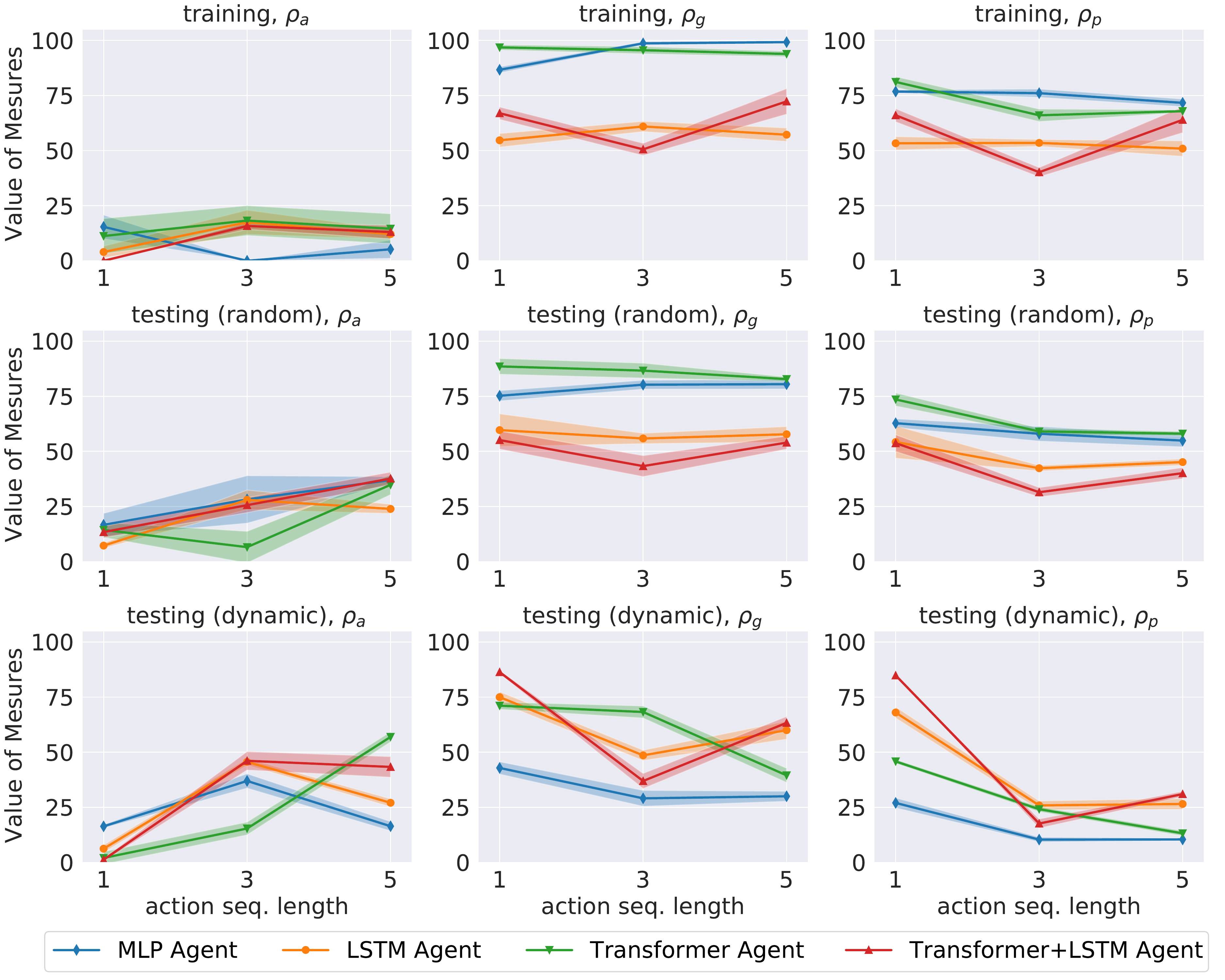}
    \caption{Ablation study of different \texttt{max\_opt\_len} (symbolic observations).}
    \label{fig:appx:abla_alen1}
\end{figure}

\begin{figure}[th!]
    \centering
    \includegraphics[width=\linewidth]{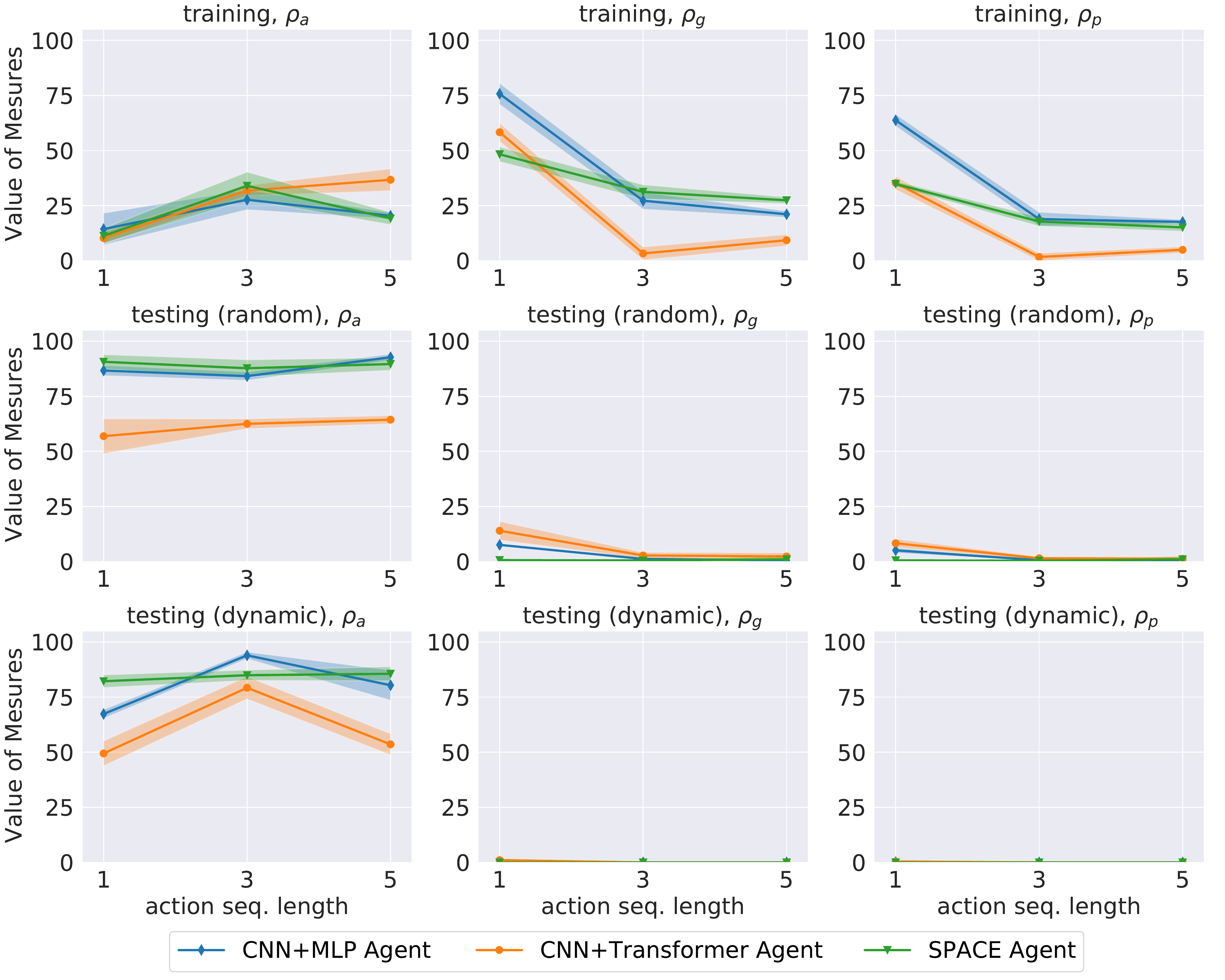}
    \caption{Ablation study of different \texttt{max\_opt\_len} (visual observations).}
    \label{fig:appx:abla_alen2}
\end{figure}

\end{document}

%% file: math_commands.tex
\usepackage{amsmath,amsfonts,bm}

\def\eqref#1{equation~\ref{#1}}
\def\1{\bm{1}}

\DeclareMathAlphabet{\mathsfit}{\encodingdefault}{\sfdefault}{m}{sl}
\SetMathAlphabet{\mathsfit}{bold}{\encodingdefault}{\sfdefault}{bx}{n}

%% file: config.tex
\usepackage{graphicx}
\usepackage{amsmath}
\usepackage{amssymb}
\usepackage{mathabx}
\usepackage{booktabs}
\usepackage{algorithmic}
\usepackage[ruled]{algorithm2e}
\usepackage{acronym}
\usepackage{enumitem}
\usepackage{hyperref}
\usepackage{balance}
\usepackage{xspace}
\usepackage{setspace}
\usepackage[compact]{titlesec}
\usepackage[skip=3pt,font=small]{subcaption}
\usepackage[skip=3pt,font=small]{caption}
\usepackage[misc]{ifsym}
\usepackage[capitalise]{cleveref}
\usepackage{tabularx}
\usepackage{multirow}
\usepackage{array}
\usepackage{makecell}
\usepackage{overpic}
\usepackage{wrapfig}
\usepackage[hang,flushmargin]{footmisc}
\usepackage{xr}
\usepackage{amsthm}
\usepackage{minitoc}

\theoremstyle{definition}
\newtheorem{definition}{Definition}

\makeatletter
\DeclareRobustCommand\onedot{\futurelet\@let@token\@onedot}
\def\@onedot{\ifx\@let@token.\else.\null\fi\xspace}
\def\eg{\emph{e.g}\onedot} 

\def\ie{\emph{i.e}\onedot}

\makeatother

\titlespacing{\section}{0pt}{1ex}{0.5ex}
\titlespacing{\subsection}{0pt}{1ex}{0.2ex}

\makeatletter
\renewcommand{\paragraph}{%
  \@startsection{paragraph}{4}%
  {\z@}{0ex \@plus 0ex \@minus 0ex}{-1em}%
  {\hskip\parindent\normalfont\normalsize\bfseries}%
}
\makeatother

\graphicspath{{figures/}}

\crefname{algocf}{alg.}{algs.}
\Crefname{algocf}{Algorithm}{Algorithms}

\definecolor{gblue}{HTML}{4285F4}
\definecolor{gred}{HTML}{DB4437}

\acrodef{halma}[HALMA]{Humanlike Abstraction Learning Meets Affordance}
\acrodef{clevr}[CLEVR]{Compositional Language and Elementary Visual Reasoning}
\acrodef{sqoop}[SQOOP]{Spatial Queries On Object Pairs}
\acrodef{scan}[SCAN]{Simplified version of the CommAI Navigation}
\acrodef{raven}[RPM]{Raven's Progressive Matrices}
\acrodef{atari}[AtariARI]{Atari Annotated RAM Interface}